%% file: main.tex
\documentclass[sigconf,authorversion,nonacm,10pt]{acmart}
\AtBeginDocument{%
  \providecommand\BibTeX{{%
    \normalfont B\kern-0.5em{\scshape i\kern-0.25em b}\kern-0.8em\TeX}}}

\setcopyright{acmcopyright}
\copyrightyear{2018}
\acmYear{2018}
\acmDOI{XXXXXXX.XXXXXXX}

\acmConference[Conference acronym 'XX]{Make sure to enter the correct
  conference title from your rights confirmation emai}{June 03--05,
  2018}{Woodstock, NY}
\acmPrice{15.00}
\acmISBN{978-1-4503-XXXX-X/18/06}

\usepackage{xcolor}
\usepackage{color}
\usepackage{graphics}
\usepackage{graphicx}
\usepackage{url}
\usepackage{multirow}
\usepackage{xspace}
\usepackage{enumitem}
\usepackage{subcaption}
\usepackage{balance}
\usepackage{acmart-taps}

\usepackage{algorithm}
\usepackage{algpseudocode}

\usepackage{xspace}
\newcommand{\name}{AIRA\xspace}

\newcommand{\remove}[1]{}

\begin{document}
\title[A Low-cost IR-based Approach Towards Autonomous Precision Drone Landing and NLOS Indoor Navigation]{\name: A Low-cost IR-based Approach Towards Autonomous Precision Drone Landing and NLOS Indoor Navigation}

\author{Yanchen Liu}
\authornote{Both authors contributed equally to this work.}
\affiliation{%
  \institution{Columbia University}
  \streetaddress{116th and Broadway}
  \city{New York}
  \state{NY}
  \country{United States}}
\email{yl4189@columbia.edu}

\author{Minghui Zhao}
\authornotemark[1]
\affiliation{%
  \institution{Columbia University}
  \streetaddress{116th and Broadway}
  \city{New York}
  \state{NY}
  \country{United States}}
\email{mz2866@columbia.edu}

\author{Kaiyuan Hou}
\affiliation{%
  \institution{Columbia University}
  \streetaddress{116th and Broadway}
  \city{New York}
  \state{NY}
  \country{United States}}
\email{kh3119@columbia.edu}

\author{Junxi Xia}
\affiliation{%
  \institution{Northwestern University}
  \streetaddress{633 Clark Street}
  \city{Evanston}
  \state{IL}
  \country{United States}}
\email{junxixia2024@u.northwestern.edu}

\author{Charlie Carver}
\affiliation{%
  \institution{Columbia University}
  \streetaddress{116th and Broadway}
  \city{New York}
  \state{NY}
  \country{United States}}
\email{cjc2306@columbia.edu}

\author{Stephen Xia}
\affiliation{%
  \institution{Northwestern University}
  \streetaddress{633 Clark Street}
  \city{Evanston}
  \state{IL}
  \country{United States}}
\email{stephen.xia@northwestern.edu}

\author{Xia Zhou}
\affiliation{%
  \institution{Columbia University}
  \streetaddress{116th and Broadway}
  \city{New York}
  \state{NY}
  \country{United States}}
\email{xia@cs.columbia.edu}

\author{Xiaofan Jiang}
\affiliation{%
  \institution{Columbia University}
  \streetaddress{116th and Broadway}
  \city{New York}
  \state{NY}
  \country{United States}}
\email{jiang@ee.columbia.edu}

\renewcommand{\shortauthors}{Liu et al.}

\input{sections/abstract}

\maketitle

\input{sections/intro}

\input{sections/related}

\input{sections/lightfield_generation}

\input{sections/ondrone_sensing}
\input{sections/nlos}

\input{sections/implementation}

\input{sections/eval}

\input{sections/eval_nlos}

\input{sections/futurework}

\section{Conclusion}
We propose \name, a light-weight, efficient, and precise method for landing and guiding drones using IR light. Unlike existing vision-based and IR-based approaches, \name requires only an off-the-shelf IR light bulb, with no modifications, at the landing station, and a small array of three PDs on the drone, costing under $83$USD. To leverage only a few PDs, we exploit the mobility of the drone to move the PDs spatially, creating a virtual array and guiding the drone towards the light source. Through our experiments, we show that \name can accurately land drones up to $11.1$ meters away, which is $9.3\times$ greater in range than traditional marker-based methods from the same height. We also demonstrate that \name can help guide and navigate drones in partial non line of sight scenarios. \name is a critical step towards low cost and low energy drone guidance for indoor environments.

\clearpage
\balance
\bibliographystyle{ACM-Reference-Format}
\bibliography{references}

\end{document}

%% file: sections/abstract.tex
\begin{abstract}
Automatic drone landing is an important step for achieving fully autonomous drones. Although there are many works that leverage GPS, video, wireless signals, and active acoustic sensing to perform precise landing, autonomous drone landing remains an unsolved challenge for palm-sized microdrones that may not be able to support the high computational requirements of vision, wireless, or active audio sensing. We propose \name, a low-cost infrared light-based platform that targets precise and efficient landing of low-resource microdrones. \name consists of an infrared light bulb at the landing station along with an energy efficient hardware photodiode (PD) sensing platform at the bottom of the drone. \name costs under $83$USD, while achieving comparable performance to existing vision-based methods at a fraction of the energy cost. \name requires only three PDs without any complex pattern recognition models to accurately land the drone, under $10$cm of error, from up to $11.1$ meters away, compared to camera-based methods that require recognizing complex markers using high resolution images with a range of only up to $1.2$ meters from the same height. Moreover, we demonstrate that \name can accurately guide drones in low light and partial non line of sight scenarios, which are difficult for traditional vision-based approaches.

\end{abstract}

%% file: sections/intro.tex
\section{Introduction} \label{sec:intro}

\begin{figure}[t]
    \centering \includegraphics[width=\linewidth]{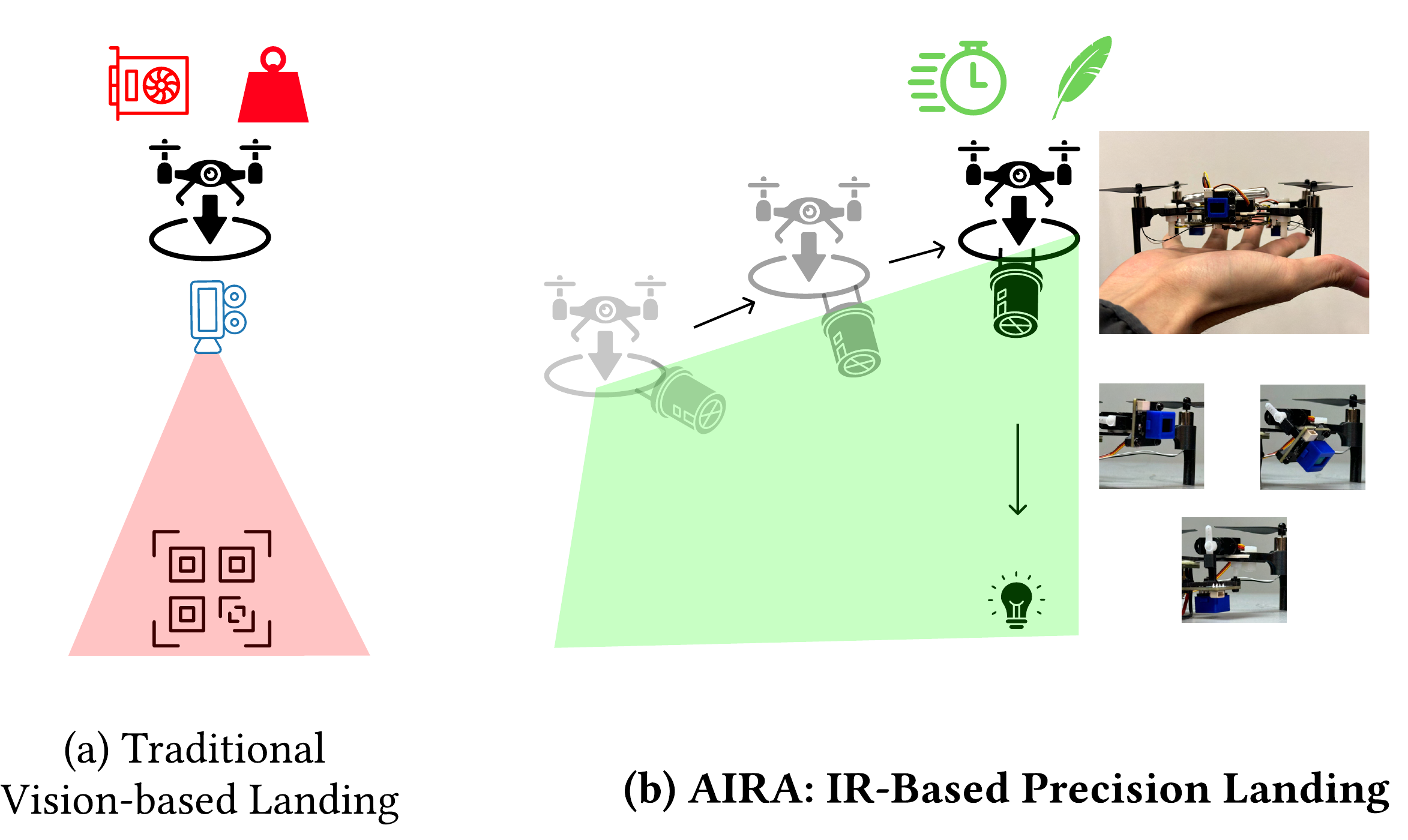}
    \caption{Overview of \name.}
    \label{fig:sys_arch}
\end{figure}

Achieving complete drone autonomy has been an active research area for decades. Among others, one key challenge is the hardware limitation and power consumption of drones. Many consumer-grade drones last tens of minutes at most, while small palm-sized drones that are more suitable for indoor usage can only fly for several minutes~\cite{minifly, droneblocks, giernacki2017crazyflie}. As such, autonomous drones need to periodically land to swap out batteries and recharge. However, automatic landing of drones is usually performed in a ``return to home'' fashion where the drone relies on its Global Positioning System (GPS) to fly back to the location it recorded at take off. This requires a larger landing space to tolerate GPS errors and is not a viable solution in indoor scenarios where GPS is obstructed. Moreover, many smaller drones such as the Crazyflie~\cite{giernacki2017crazyflie} do not have onboard GPS and need to land manually.

While there are works that target drone landing with other sensing modalities, such as through visual markers, wireless anchor points, and acoustic-based approaches, most of these platforms are targeted for drones with enough compute, battery life, and carrying capacity to support image/RF processing and active sensing. These drones tend to be on the larger side, which are not conducive for indoor applications where space for safe navigation is constrained (e.g., Holybro X500 weighs hundreds of grams with tens of centimeters in length, width, and height~\cite{holybrox500}).

Leveraging infrared (IR) light to guide and land drones is advantageous over camera-based methods because it lies outside the visible spectrum and is not affected by ambient light conditions (e.g., at night). Existing IR-based guidance and landing systems generally use IR beacons along with an IR camera to perform pattern recognition. Like RGB camera-based systems, this requires comparatively expensive cameras. We demonstrate how drone landing and guidance can be achieved with just a small array of photodiodes (PD), rather than hundreds of thousands of channels as is common in camera-based methods, without the need to perform expensive pattern recognition to localize markers.

We propose \name, \underline{a} low-cost, energy-efficient, infrared (\underline{IR}) light-based platform for \underline{a}utonomous and precise drone landing (Figure~\ref{fig:sys_arch}). Unlike existing works, \name targets microdrones that are often smaller than palm-size with low battery life to support active or high resolution sensing. \name consists of two simple hardware components: 1) a landing station with an off-the-shelf IR light bulb with 2) a small 3 photodiode array on the drone to guide it to land, even in partially non line of sight scenarios. Unlike camera-based methods that require hundreds of thousands of pixels to generate images and perform pattern recognition, \name localizes the direction to guide the drone by exploiting the drone's mobility to create a virtual PD array, allowing a single PD to sense light intensity in multiple directions and converge on the landing station. As such, \name does not require any complex pattern recognition, other than sensing light intensities, or methods to account for multipath as is common in acoustic and RF-based methods. Additionally, \name requires no modifications or complex modulation schemes to the IR light source.

Our contributions are as follows:

\noindent
\textbf{Low-cost, feather-weight, IR-based drone landing platform}: We propose \name an end-to-end hardware and software platform for precise drone landing that targets low-resource palm-sized microdrones. Unlike existing works that rely on active audio sensing, high resolution imaging, or RF sensing that small microdrones often cannot sustain due to power or payload constraints, \name only leverages a small array of 3 PDs, with an IR light source at the landing station, to guide itself. In total, \name costs under $83$USD, while weighing less than $18$g, allowing even microdrones with tens of grams of payload to carry (e.g., Crazyflie).

\noindent
\textbf{Efficient Localization and Guidance Methods Exploiting the Drone's Mobility.} We propose novel methods to guide the drone to the landing station by exploiting the drone's mobility to onboard PD's spatially. Unlike existing vision-based works that require expensive pattern recognition and high fidelity images, \name simply guides the drone to the landing station by following the direction with the greatest light intensity that can be measured with only several PDs.

\noindent
\textbf{Demonstration in real and partially non line of sight settings.} We deploy \name in a variety of realistic indoor environments, and demonstrate \textit{similar landing performance} compared to existing vision-based approaches, up to $1.5$m away and $1.2$m above ground, at a \textit{fraction of the energy cost} ($\mu W$ level vs $mW$ level for vision-based methods), from up to $11.1$ meters from the landing station. \textit{\name's range is greater than $9.9$ meters compared to vision-based approaches.} Moreover, we demonstrate successful guidance in several non line of sight scenarios where the drone begins at a position that is occluded from the light source.

%% file: sections/related.tex
\section{Related Work} \label{sec:related}
\noindent\textbf{Drone Localization}. Besides using traditional GPS / GPS-RTK based methods, researchers have explored RF methods to localize drones. However, most of these localization methods suffer from accuracy issues, with 3-D localization errors at best in 10s of centimeters. Additionally, GPS-based systems see limited accuracy in indoor scenarios and other types of wireless localization schemes (e.g., WiFi, ultra-wideband (UWB), active acoustic sensing)~\cite{bisio2021localization, dhekne2019trackio, mao2017indoor, chi2022wi, sun2022aim, famili2023idrop} have limited operation range and require additional hardware support that palm-sized drones often cannot support computationally. Works that leverage vision to localize a drone often have limited field of view and see performance degradation in low-light conditions~\cite{pavliv2021tracking, mraz2020using}. Additionally, leveraging passive audio sensing to localize drones often suffers from the interfering noise of the propellers~\cite{manamperi2022drone, chen2021pod}.

\noindent\textbf{Localization for Drone Landing}. However, when the application requirements shifts from general localization (needing to know (x, y, z)), into moving the drone onto a specific location, precise localization for drones becomes easier. To guide the drone to a landing target, researchers have leveraged markers, beacons, and anchors with varying modalities including UWB~\cite{ochoa2022uwb, zeng2023uav} and visual objects and markers~\cite{kim2021roland}. Placing a visual marker at the landing station reduces the landing problem to needing to estimate the relative location of the drone to the landing target~\cite{grlj2022decade}; as such, the most commonly used method for autonomously landing drones is to use RGB cameras on the drone to locate a pattern, usually a QR code, placed on the target landing location \cite{wang2020precision, nguyen2018lightdenseyolo, chen2016system}.These systems achieve high accuracy and have been deployed in commercial systems, such as in Google's project wing \cite{projectwing} for package deliveries. However, despite the high accuracy, camera-based approaches running computer vision algorithms incurs a computational cost often not supported by small microdrones, and methods that leverage visible light often experience degraded performance in low-light conditions (e.g., at night) and cluttered scenarios (e.g., in an office or home setting).

\cite{wang2022micnest, he2023acoustic} propose approaches that emit acoustic pulses to localize and guide drones. While this class of methods can overcome reduced performance observed by vision-based approaches in low-light scenarios, emitting a signal in active sensing approaches requires additional payload and reosurces beyond what a typical microdrone can provide.

\noindent
\textbf{Infrared Methods.} There are a number of works that introduce IR-based methods for localization~. Works that leverage IR to guide and land drones typically leverage IR tags~\cite{springer2024precision,kalinov2019high,khithov2017toward} or LED matrices~\cite{kong2013autonomous, yang2016ground,janousek2018precision,nowak2017development} to create patterns that the drone can detect. These methods still require the use of a full IR camera, often in conjunction with other sensors such as RGB cameras, lidar, and IMU to be effective. This incurs heavy sampling, compute, and price cost. On the contrary, our work focuses on drone landing and guidance with only IR light. Additionally, we reduce the number of sensor channels from hundreds of thousands pixels to three or less, while providing accurate guidance without complex pattern recognition.

%% file: sections/lightfield_generation.tex
\section{Infrared Light Field Generation} \label{sec:lightfield_generation}

In this work, we target future small drone platforms that may only have the payload and computation to support several PDs, rather than a full camera. A camera system for landing a drone could consist of placing a QR code at the landing site and using the camera to detect the pattern and estimate its position. To reduce computation, can we reduce the resolution of the camera to a single pixel and remove pattern recognition? If you were asked to walk to a bright spot in an empty dark room while being blind-folded, what would you do? You could take one step forward, and see whether or not you see more lights through the blindfold. If you are walking in the right direction, you would see more lights gradually, and inversely if you are seeing less light, you are walking in the wrong direction and should try a different direction. In our problem, the ``bright spot'' is the ground station generating an IR light field, while the ``blind-folded'' person is our light-weight drone equipped with a ``one-pixel camera'' -- a photodiode.

In this section, we explore different types of infrared light fields that can be generated and deployed at the landing station and analyze their impact on landing performance. In Sections~\ref{sec:lightfield_sensing_on_drone} and~\ref{sec:explore_nlos}, we explore different configurations for sensing the light field on drones.

\subsection{Generating Light Fields}
\label{subsec:generating_light_fields}

\begin{figure}
    \centering
    \includegraphics[width=\linewidth]{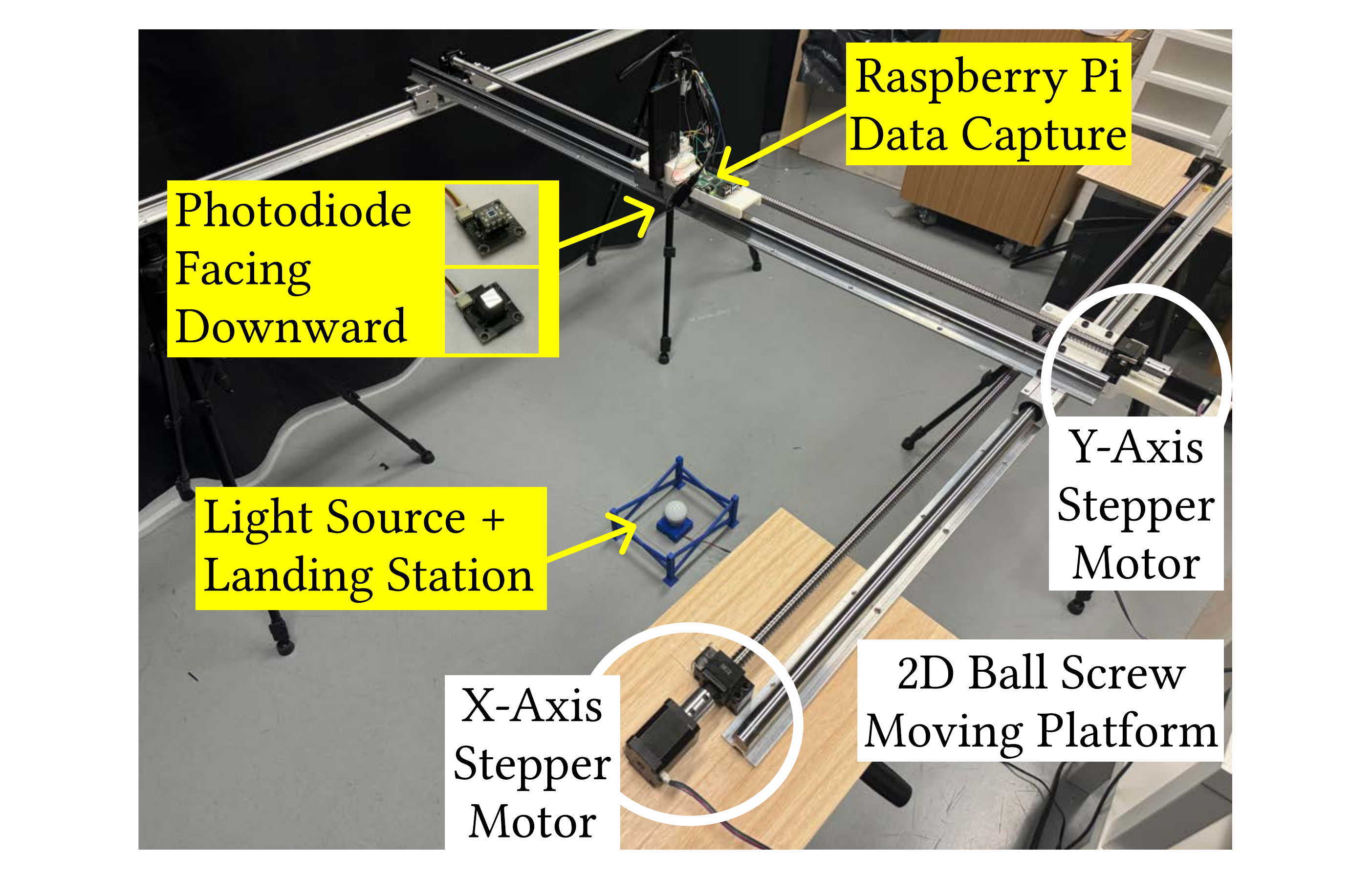}
    \caption{Setup for measuring light fields.}
    \label{fig:lightfield_measurement_setup}
\end{figure}

\begin{figure*}[t]
    \centering \includegraphics[width=\linewidth]{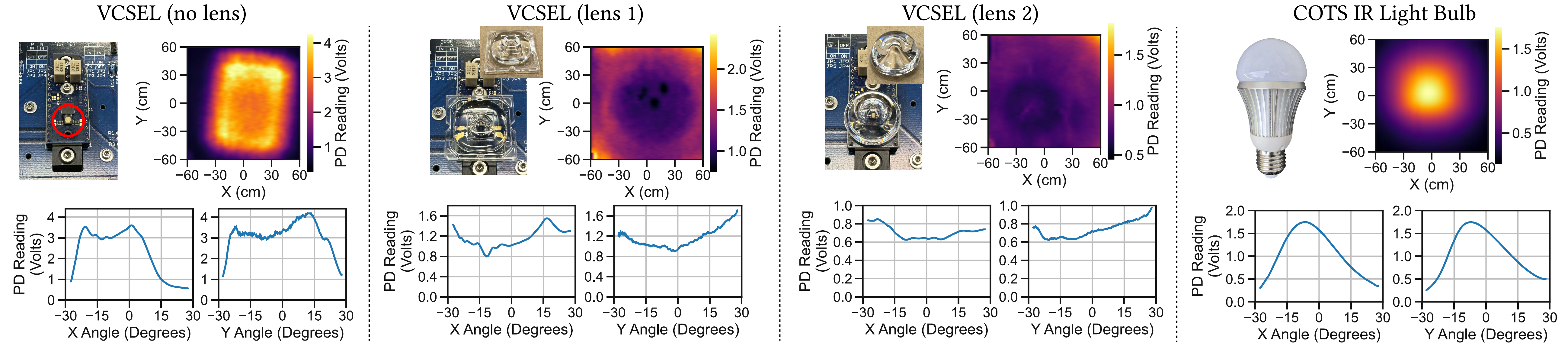}
    \caption{Measured lightfields using off-the-shelf components.}
    \label{fig:measured_lightfields}
\end{figure*}

\subsection{Light Field Selection}
\label{subsec:lightfield_selection}

\begin{figure*}[t]
    \centering
    
    \begin{subfigure}[b]{0.22\textwidth}
        \includegraphics[width=\textwidth]{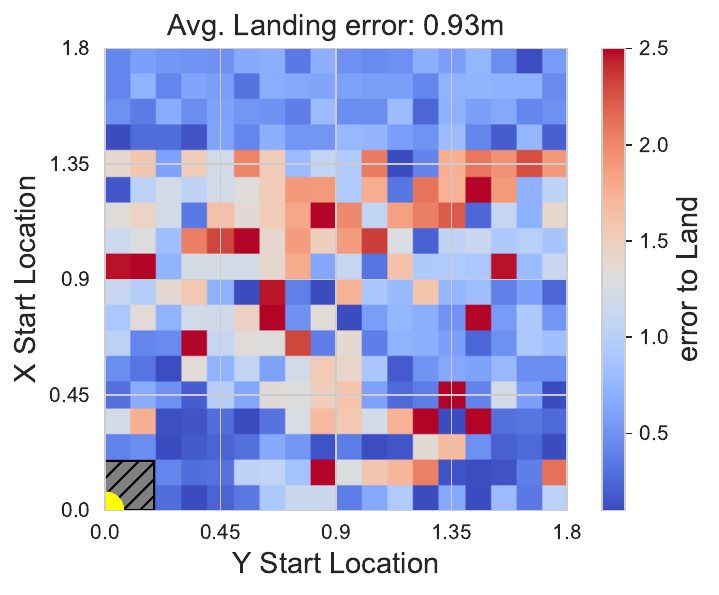}
    \end{subfigure}
    \hfill
    \begin{subfigure}[b]{0.22\textwidth}
        \includegraphics[width=\textwidth]{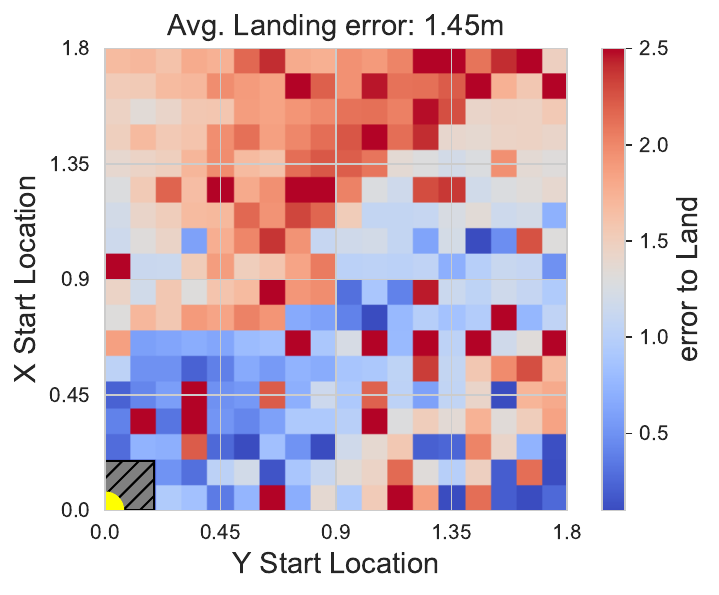}
    \end{subfigure}
    \hfill
    \begin{subfigure}[b]{0.22\textwidth}
        \includegraphics[width=\textwidth]{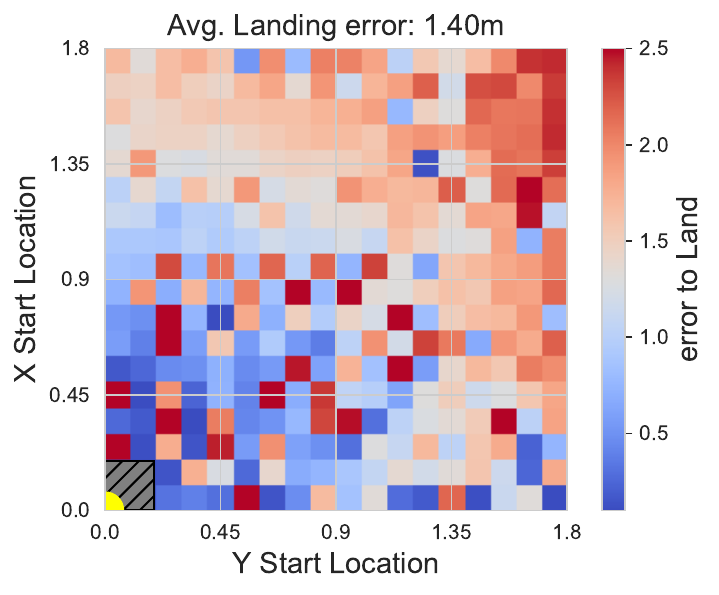}
    \end{subfigure}
    \hfill
    \begin{subfigure}[b]{0.22\textwidth}
        \includegraphics[width=\textwidth]{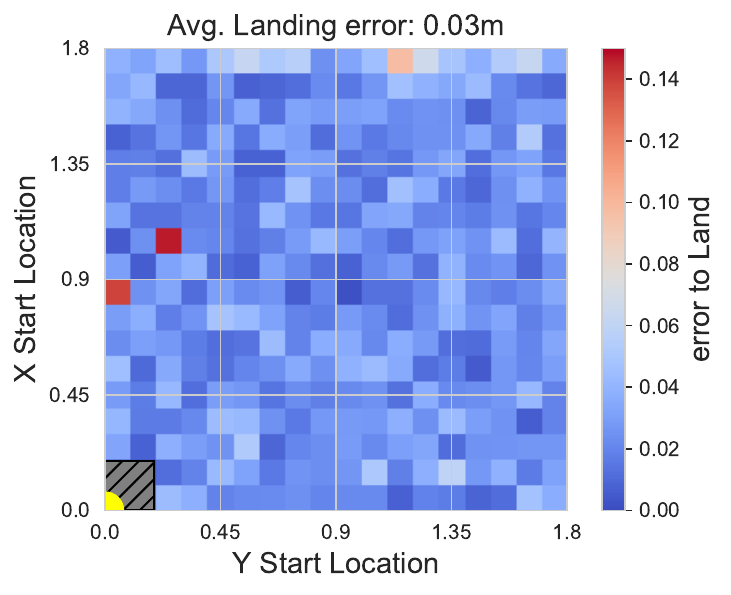}
    \end{subfigure}

    \begin{subfigure}[b]{0.22\textwidth}
        \includegraphics[width=\textwidth]{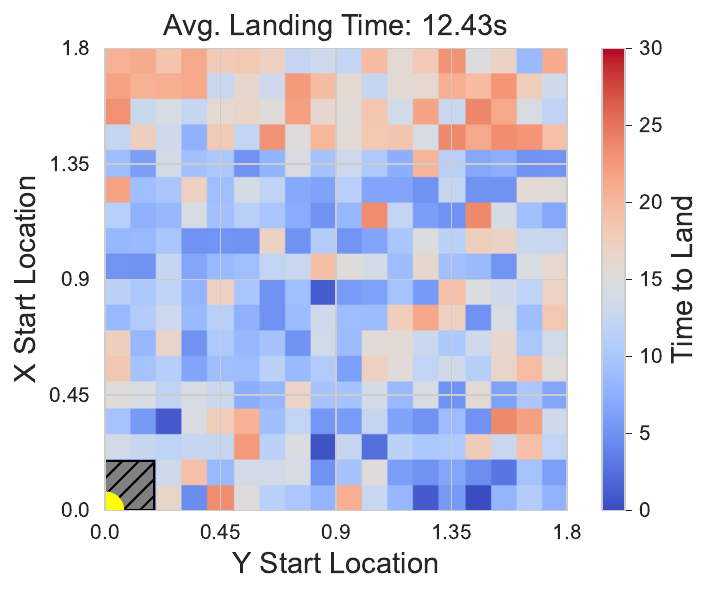}
    \end{subfigure}
    \hfill
    \begin{subfigure}[b]{0.22\textwidth}
        \includegraphics[width=\textwidth]{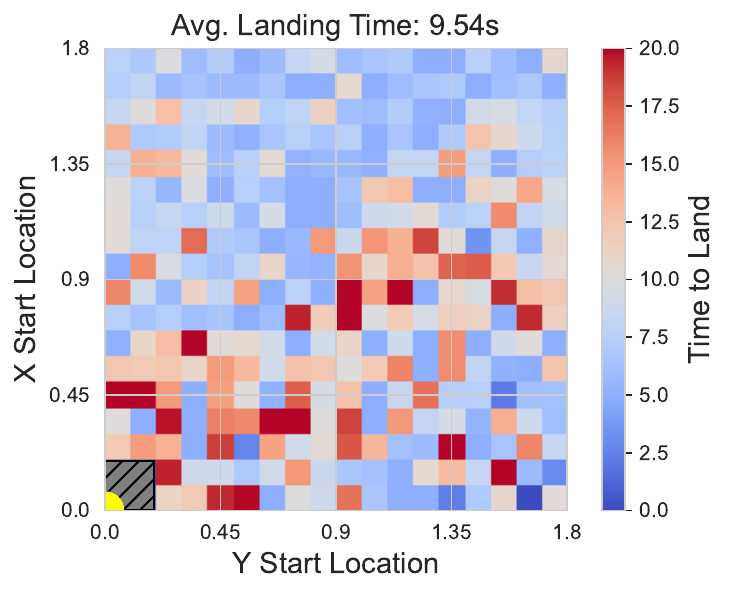}
    \end{subfigure}
    \hfill
    \begin{subfigure}[b]{0.22\textwidth}
        \includegraphics[width=\textwidth]{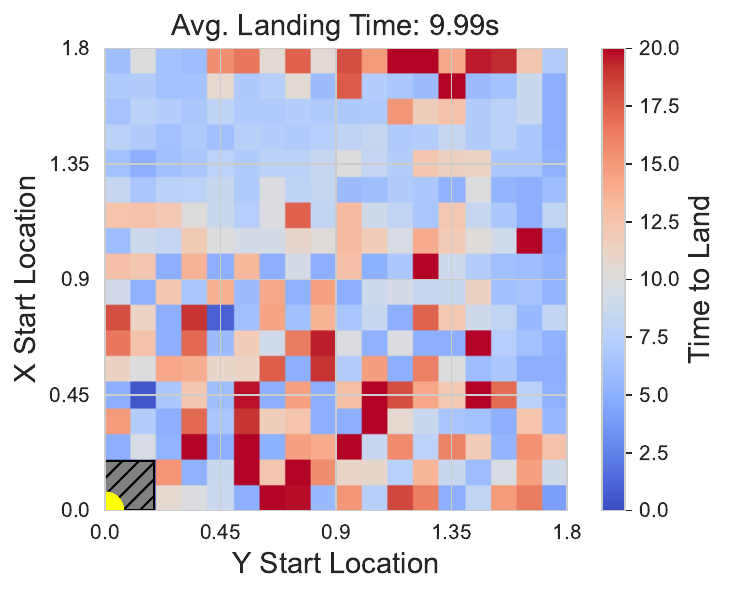}
    \end{subfigure}
    \hfill
    \begin{subfigure}[b]{0.22\textwidth}
        \includegraphics[width=\textwidth]{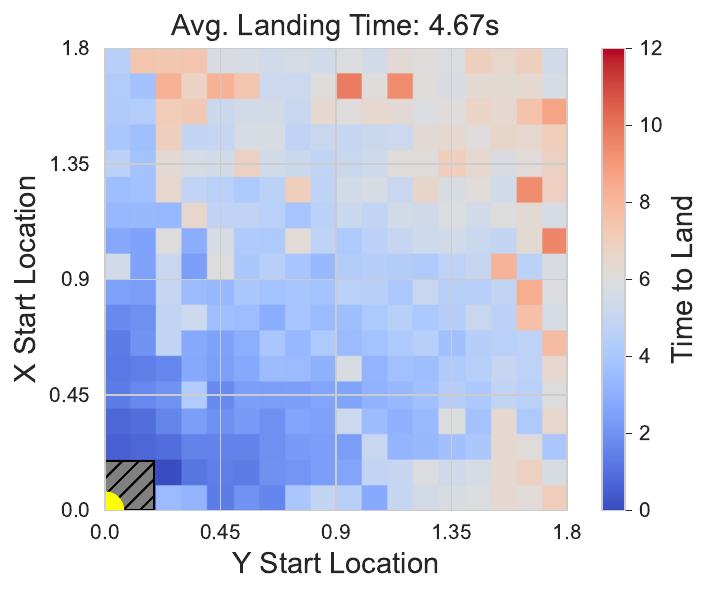}
    \end{subfigure}

    \begin{subfigure}[b]{0.22\textwidth}
        \includegraphics[width=\textwidth]{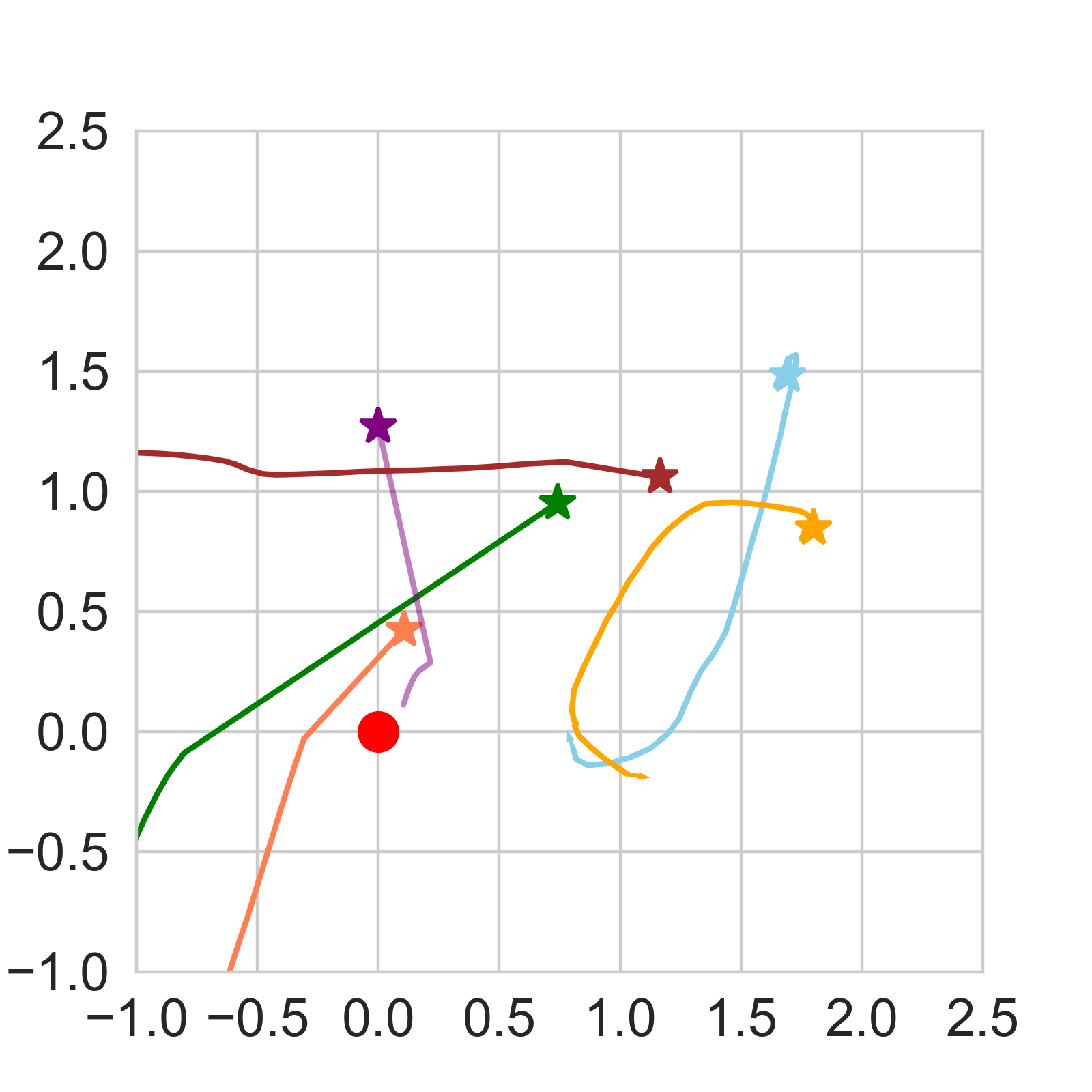}
        \caption{No Lens}
        \label{fig:no_lens_sim}
    \end{subfigure}
    \hfill
    \begin{subfigure}[b]{0.22\textwidth}
        \includegraphics[width=\textwidth]{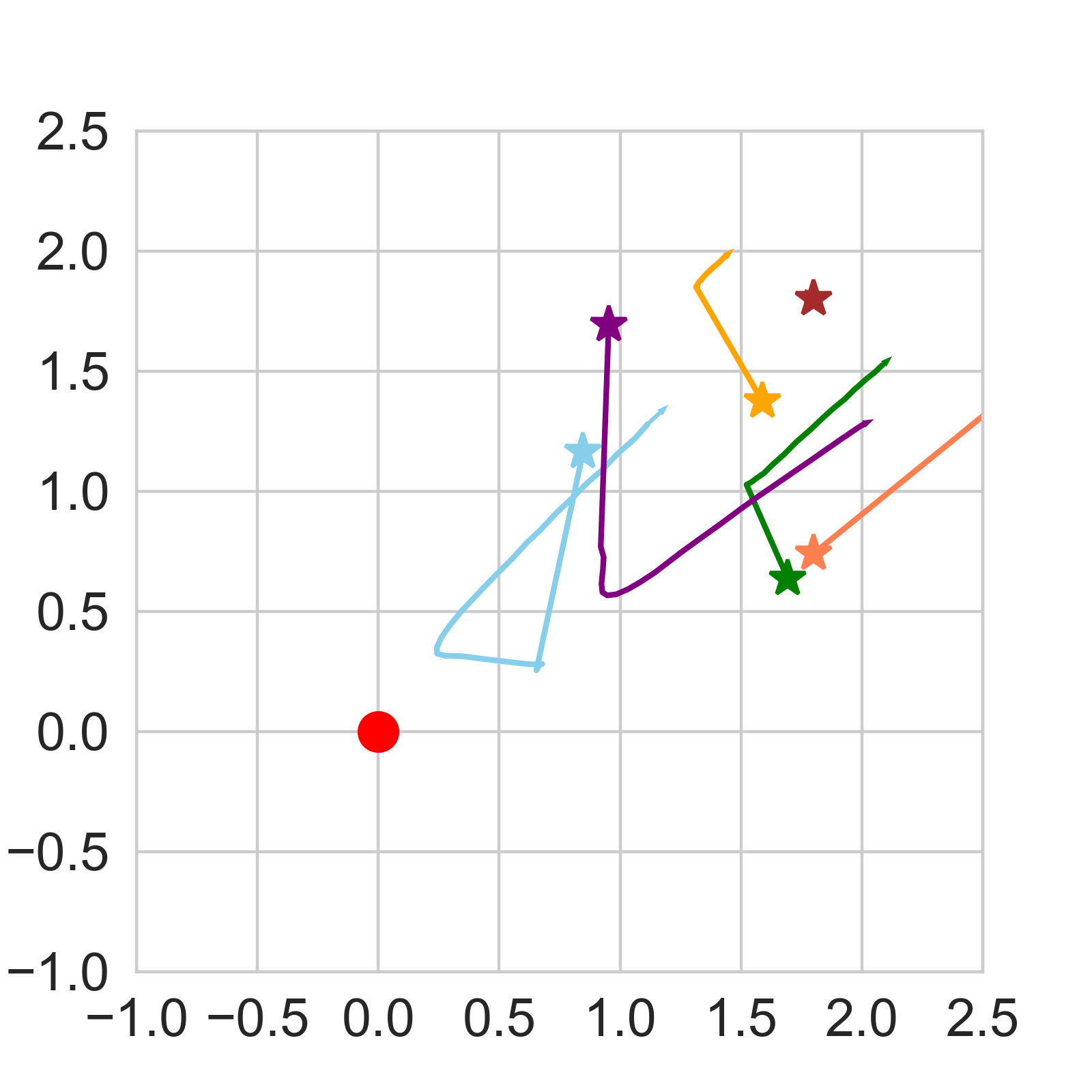}
        \caption{Lens 1}
        \label{fig:lens1_sim}
    \end{subfigure}
    \hfill
    \begin{subfigure}[b]{0.22\textwidth}
        \includegraphics[width=\textwidth]{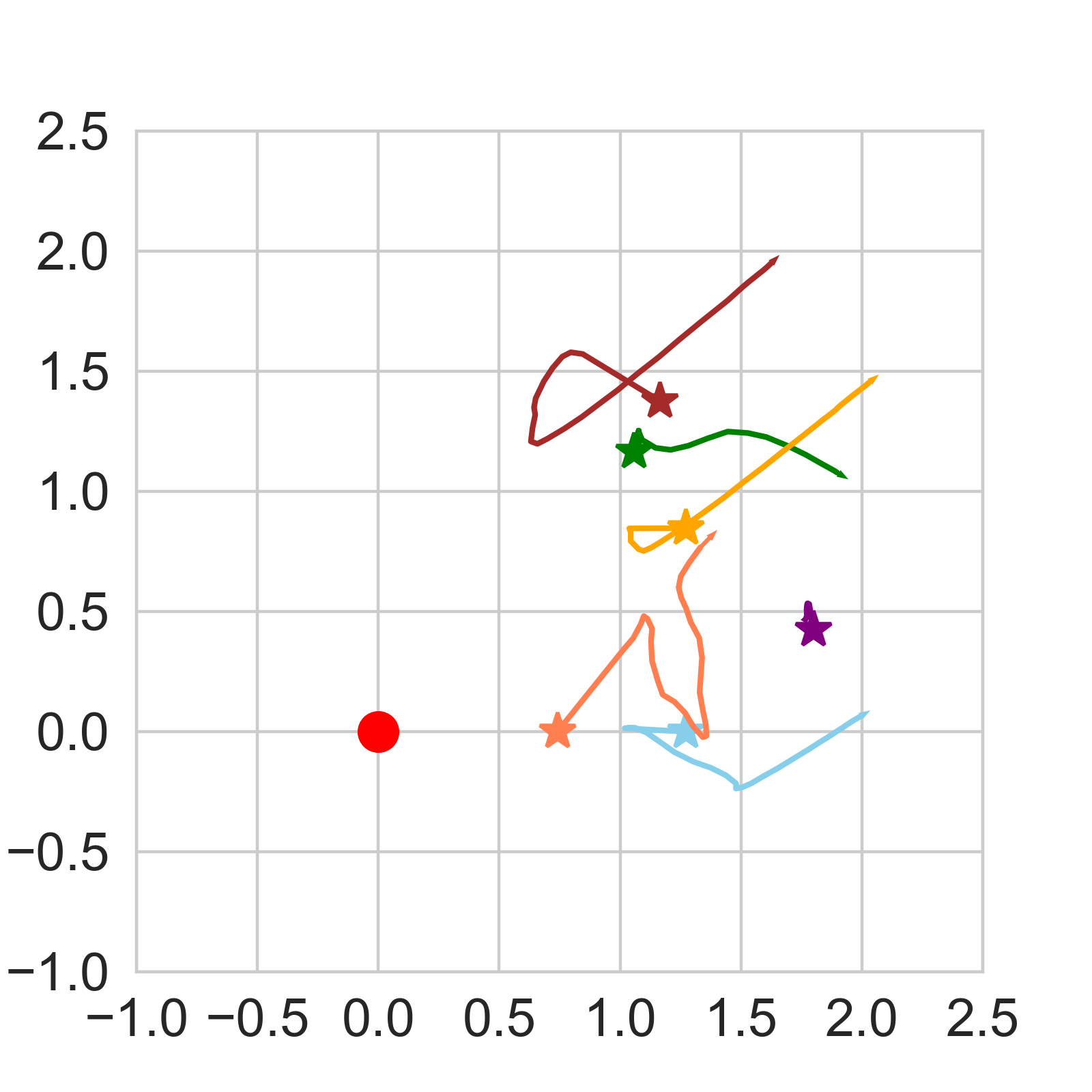}
        \caption{Lens 2}
        \label{fig:lens2_sim}
    \end{subfigure}
    \hfill
    \begin{subfigure}[b]{0.22\textwidth}
        \includegraphics[width=\textwidth]{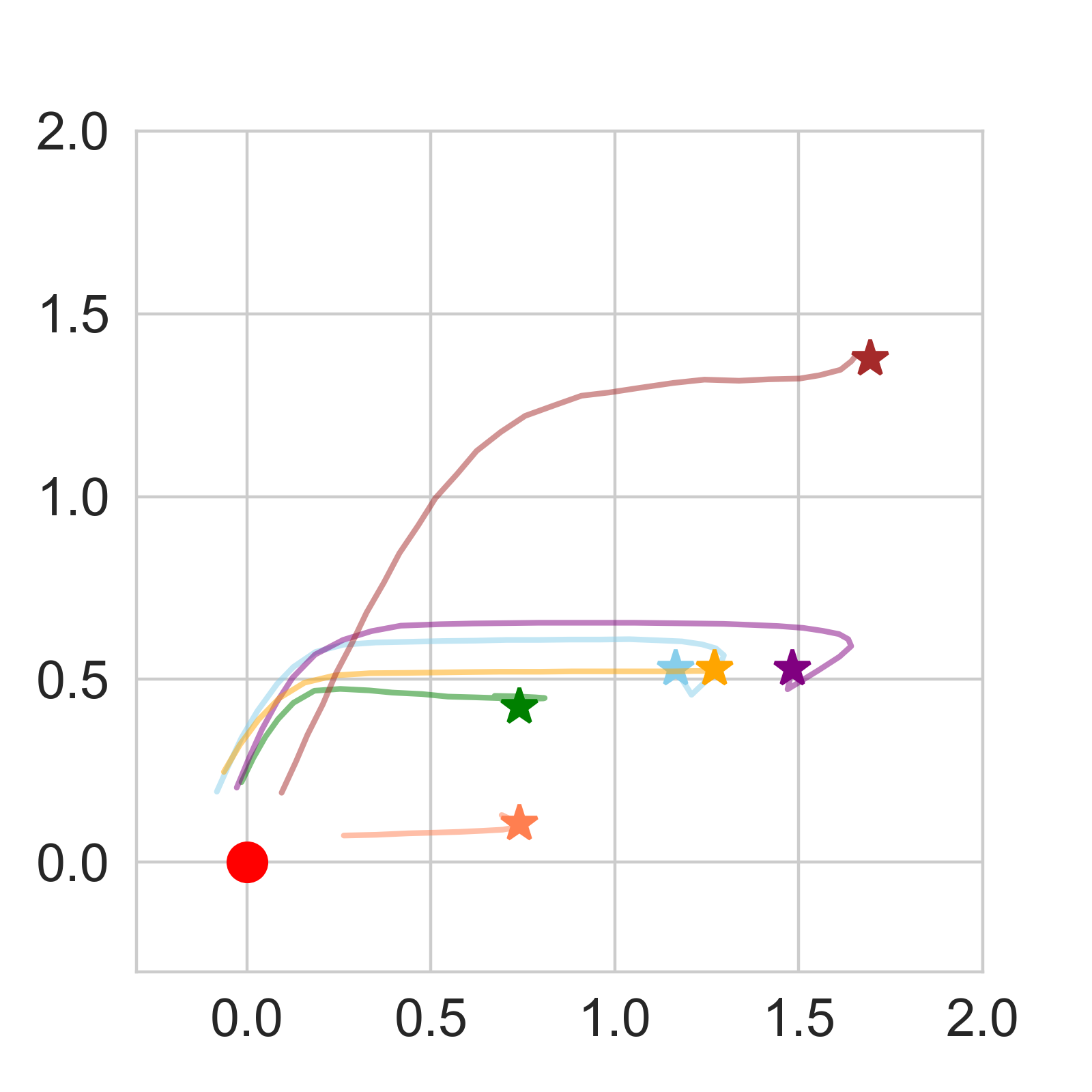}
        \caption{Light Bulb}
        \label{fig:bulb_sim}
    \end{subfigure}
    \caption{Comparing light field landing time, offset, and example trajectories across four methods of generating light fields.}
    \label{fig:lightfield_simulation_comparison}
\end{figure*}

To measure potential light fields, we created the setup shown in Figure~\ref{fig:lightfield_measurement_setup}, where we placed the light source in the middle of a frame made up of two 1.2m camera sliders in a 2D grid. We attach a downward-facing PD with a Raspberry Pi~\cite{rpi} to collect light intensities along with an HTC Vive VR controller~\cite{htcvive} to obtain measurement locations. To measure the light field at different heights, we change the height of the light source from the ground.

Figure~\ref{fig:measured_lightfields} shows four different infrared light sources we characterized, including slices or cross sections along the $X$ and $Y$ axes. The light fields in Figures~\ref{fig:measured_lightfields}a-c were generated with a vertical-cavity surface-emitting laser (VCSEL) and placing diffuser lenses over them. The light field in Figure~\ref{fig:measured_lightfields}d was measured from an off-the-shelf IR light bulb. We see that all of the light fields generated have distinct patterns, but the IR light bulb (arguably the simplest solution) generates a field that is concave and centered around the light source (target landing site), similar to a Gaussian curve. This concavity, in theory, enables a very simple paradigm for determining the direction of the landing target. which direction to move: towards the direction where the light intensity increases. All other methods require the drone to make correlations between a map of the light field and observed intensities to estimate its location within the landing area; the drone needs to store the radiation pattern of the light field in memory and compute correlations, much like fingerprinting methods used in many localization works that are computationally and memory intensive~\cite{jun2017low}. We conduct an initial exploration of the landing performance that these light fields provide in conjunction with the PD guidance methods next.

\subsection{Drone and Light Field Simulation Environment}
\label{subsec:drone_lightfield_sim}

Using the light fields measured in Section~\ref{subsec:generating_light_fields}, we built a drone landing simulator in Gazebo~\cite{koenig2004design}, a widely used robotics simulator, to analyze the impact each light field has on landing performance.

Figure~\ref{fig:lightfield_simulation_comparison} shows heat maps of landing performance across different light fields at a starting height of 1.1m, as well as example landing trajectories. Since all of the fields are symmetric across quadrants, we show only one quadrant (upper right) for simplicity. Each pixel indicates the starting point of the drone, while the color indicates either the offset of the landed drone from the center of station or the time required to make the landing. On the drone, we simulated an array of 6 PDs used to guide the drone to the landing station using methods described in Section~\ref{subsec:sensing_configurations} (ArPD6).

We see that there is high landing error for each VCSEL + lens generated light field (light fields 1 through 3). Light field 1 (no lens) generated a bimodal radiation pattern rather than a concave pattern with a single peak. Light field 2 and 3 (lens 1 and 2) generated a similar pattern, except the peaks are diffused further out from the light source, and the landing station happens to center around a local minimum. For each of these three light fields, the starting position of the drone significantly affected where the drone landed; the drone often climbed to one of these off-centered peaks. The IR light bulb (light field 4) had low and relatively uniform error no matter the starting position because of its concave radiation pattern; the drone always looks towards the direction of greatest brightness. Moreover, this simple pattern allowed the drone to land faster than any other method (around 5 seconds vs. 10 seconds). As such, \textbf{we use the IR light bulb to generate the light field at the landing station} in \name.

%% file: sections/ondrone_sensing.tex
\section{Light Field Sensing on Drones} \label{sec:lightfield_sensing_on_drone}

Continuing from our ``blind-folded person'' analogy, there are intuitively two potential (related) configurations for sensing and guiding the drone towards the light source, which we introduce and analyze next. An illustration of these is shown in Figure~\ref{fig:pd_patterns}. We focus on line of sight (LOS) scenarios in this section and discuss non line of sight cases (NLOS) in Section~\ref{sec:explore_nlos}.

\subsection{Photodiode Sensing Configurations}
\label{subsec:sensing_configurations}

\begin{figure*}[t!]
    \centering
    \includegraphics[width=0.9\linewidth]{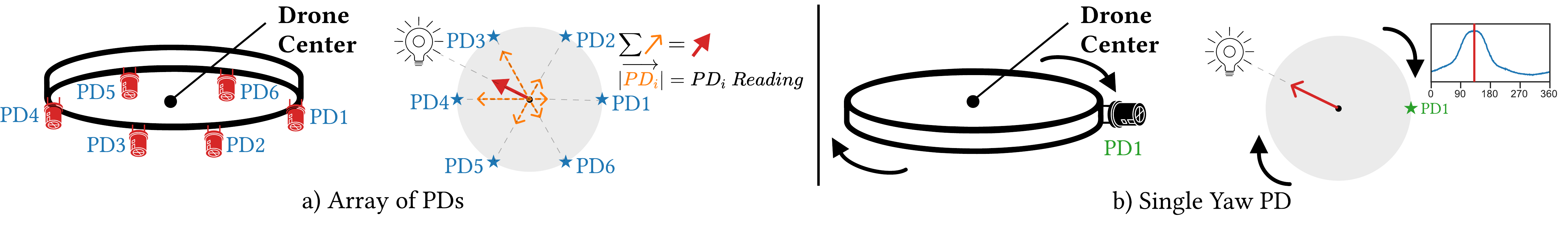}
    \caption{PD arrays, methods, and placements considered. a) array of PDs (6 shown) - ArPD6 and b) single PD where the drone sweeps 360 degrees along its yaw.}
    \label{fig:pd_patterns}
\end{figure*}

\noindent
\textbf{1. Array of PDs (ArPD).} Placing an array of PDs on the drone, that are spread out, allows the drone to leverage the spatial diversity of the sensors to localize the direction of the light source. Figure~\ref{fig:pd_patterns}a shows an array of 6 PDs (ArPD 6). As an example, if the light source is on the left side of the drone, the PDs on the left side (PD3, 4 and 5) will likely see greater intensity, which will be the direction that the drone moves.

To determine the direction that the drone moves, we treat the location of each PD as vectors with respect to a reference point (e.g., the center of the drone). This captures the directionality of each PD on the drone. The light intensity measurement at each PD is used to scale each vector before summing, which results in a single vector that points in the direction that the drone moves. %

\noindent
\textbf{2. Single PD (SPD).} If we reduce an array of PDs down to just a single PD, then directionality is lost and it is not possible to estimate the direction of the light source. However, unlike static arrays, the drone is mobile and capable of moving and turning. By placing a single off-centered PD, the drone can create a virtual array by turning and sweeping 360 degrees along its yaw. The drone can then move in the direction of the highest intensity.

While using a single PD can decrease cost, power consumption, and has a smaller required payload compared to using an array, the drone is required to spend time turning, which can significantly increase landing time. Moreover, operating a small drone with precision to make small movements is very challenging. This difficulty arises from the drone's intricate control systems, which continually adjust in response to sensor feedback, which lead to slight oscillations or corrections as the drone compensates for variations in motor mechanics, external airflow disturbances, and other factors. Sensor errors and drift also contribute to these challenges. While these minor deviations might not immediately impact the system's performance, they are not easily correctable and can significantly affect the drone's landing accuracy, particularly when the drone is close to the landing station and about to land. Next, we analyze the performance of these sensing configurations depending on the number and angle of the PDs.

\subsection{Number of Photodiodes}
\label{subsec:num_pd}

\begin{figure*}
    \centering
    \includegraphics[width=0.95\linewidth]{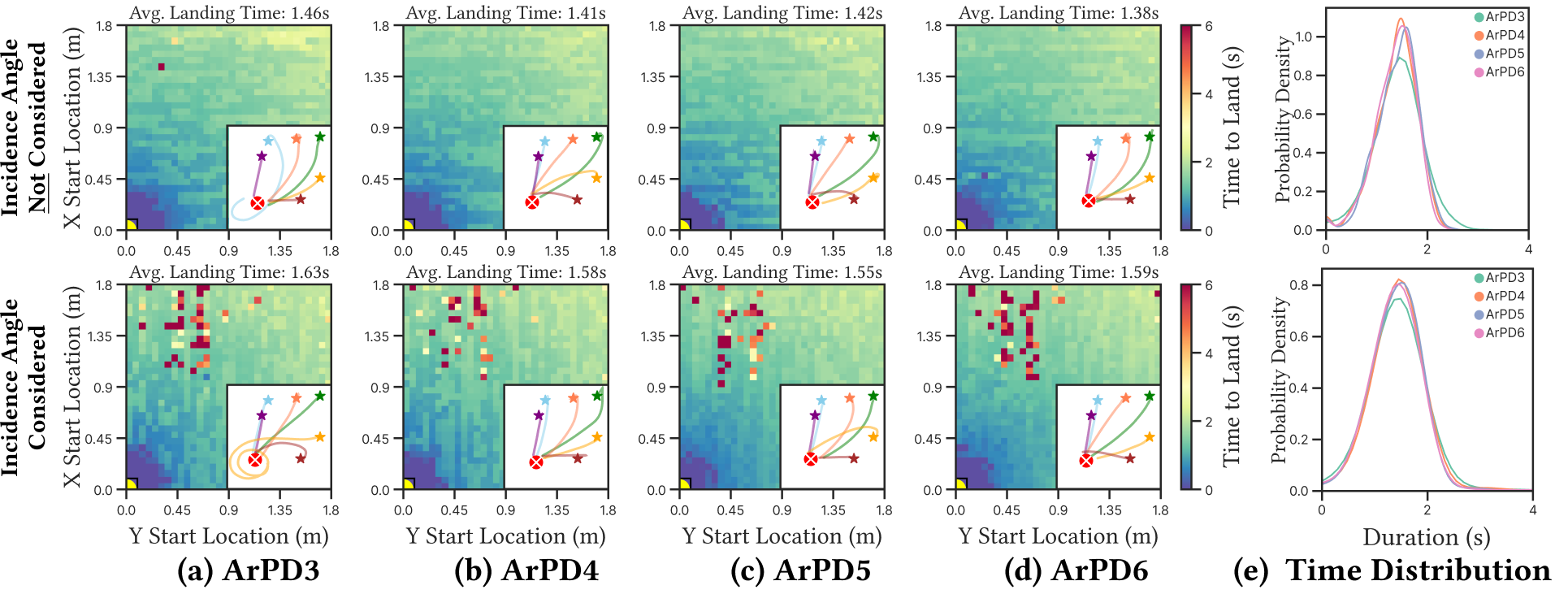}
    \caption{Number of PDs vs landing time with example trajectories.}
    \label{fig:simulation_num_pd_time_and_example_path}
\end{figure*}

Here, we analyze the impact of the number of PDs on the landing performance using our drone and light field simulator (Section~\ref{subsec:drone_lightfield_sim}). Figure~\ref{fig:simulation_num_pd_time_and_example_path} shows a heat map of the landing error and time from different starting points in the x-y plane at a height of 1.1m, as well as several example trajectories, while varying the number of PDs in the array. The average landing error was fairly consistent even using just 3 PDs compared to many more (e.g., 16). While the average landing time slightly increased as the number of PDs was reduced, the difference between using three PDs versus 16 is less than $3$ seconds (approximately $7$ seconds with 3 PDs and $4$ seconds with 16), which is almost negligible. Figure~\ref{fig:simulation_num_pd_time_and_example_path}e shows the distribution of landing times with varying number of PDs; the distributions of each quantity of PDs are all very similar. This suggests that we can aggressively reduce the number of PDs if we employ an array.

\subsection{Angle of Photodiodes and Operating Range}
\label{subsec:angle_pd}

\begin{figure}
    \centering
    \includegraphics[width=\linewidth]{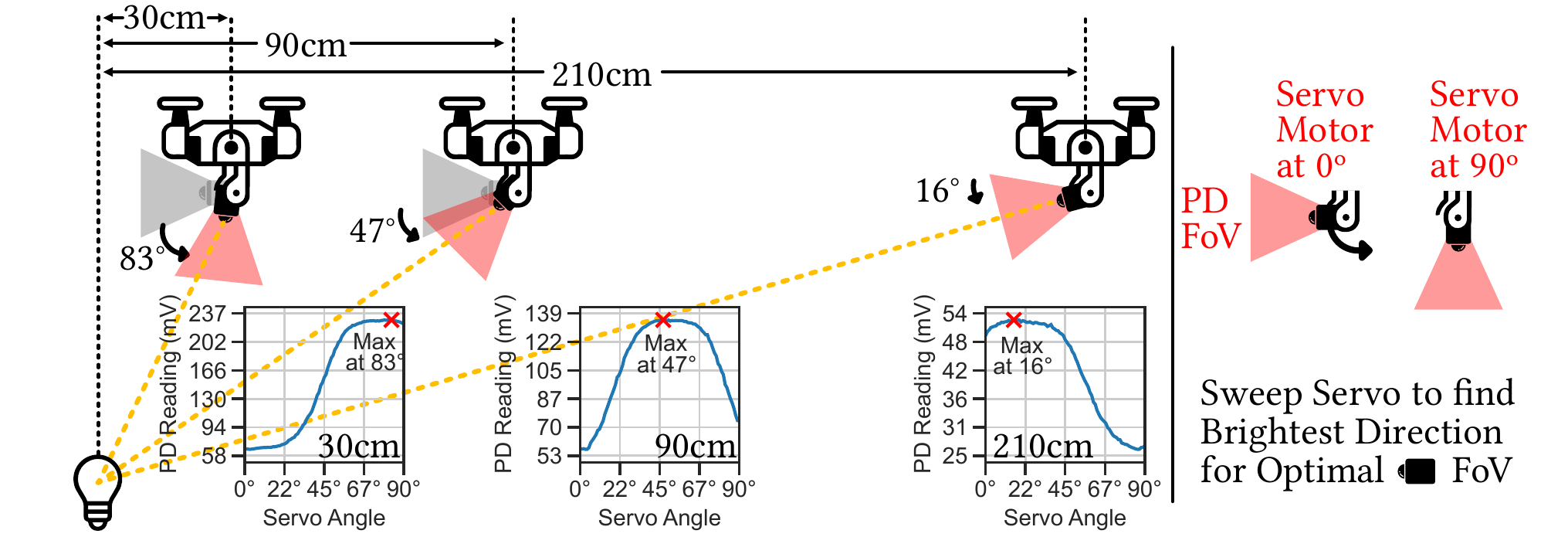}
    \caption{Incident angle of light depends both on the angle of the PD on the drone and the distance from the landing station. High incident angle results in low intensity readings. Because of these variations, we incorporate a motorized PD that allows us to change the angle of the PD depending.}
    \label{fig:impl_incident_light}
\end{figure}

In the previous measurements and simulations, we assumed that the PDs are facing downwards. However, the readings obtained from the PD depend significantly on the angle of the incident light on the sensor. This depends on 1) the distance away from the light source and 2) the angle of the PD on the drone, as illustrated in Figure~\ref{fig:impl_incident_light}. We take measurements at multiple distances from the light source, at a height of $1$ meters, and sweep the angle of the PD from $0$ degrees (side facing) to $90$ degrees (downward facing). When the drone is far from the light source (> $1.5$ meters), the PD sees the most light intensity when it is side-facing ($0$ degrees), since the incident angle of light on the sensor is smallest (e.g., the light is hitting the PD directly). However, as we move closer to the drone, we see that this peak shifts to a greater angle until the drone is within around $1$ meters. Within this range, the PD is completely downward facing ($90$ degrees) and measures the greatest light intensity. As we will see in Section~\ref{sec:eval_yanchen}, the angle of the PD greatly impacts the operating distance or range from the landing station where \name can reliably guide the drone.

\subsection{Proposed Motorized PD Approach}
\label{subsec:proposed_motorized_spd}

\begin{figure}
    \centering
    \includegraphics[width=0.9\linewidth]{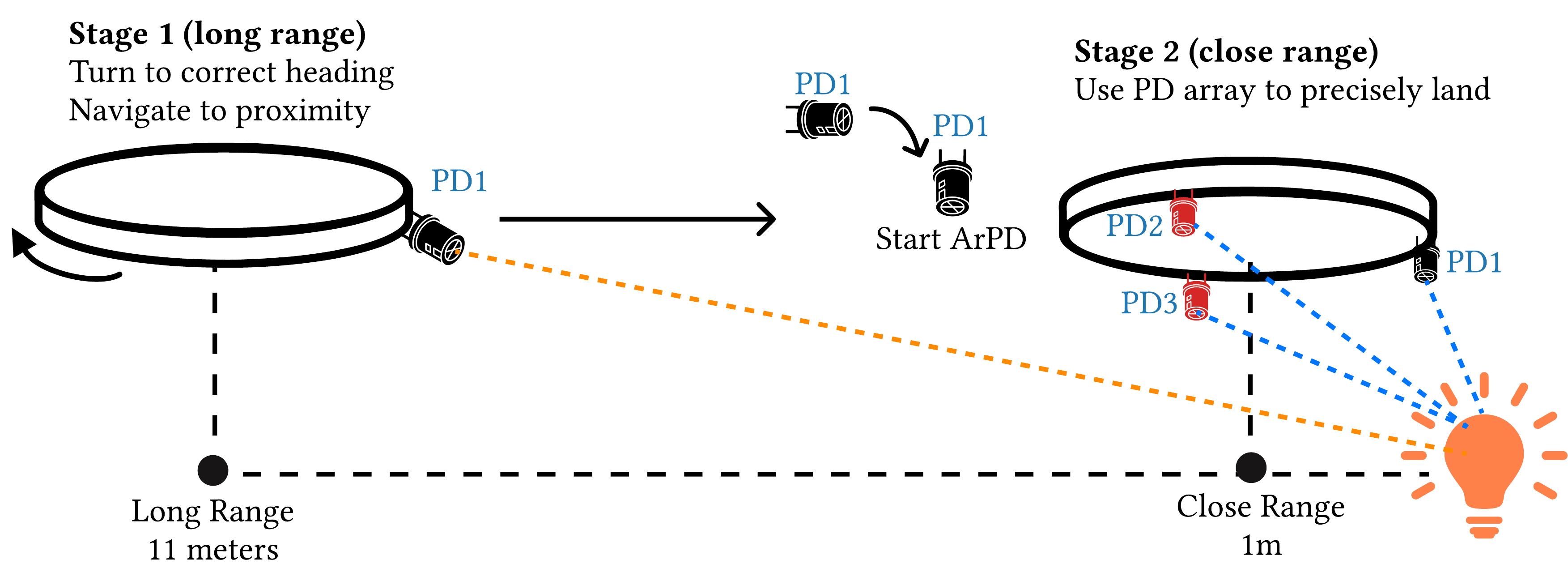}
    \caption{Proposed landing method combining a motorized PD along with two downward facing PDs. When far from the landing station (> $1$ meters) \name uses the single motorized PD to guide the drone towards the landing station, while actuating the PD to the polar angle with the greatest intensity. When the drone is close to the landing station, the downward facing PDs measure greater light intensity and the \name switches to using all three PDs as an array (ArPD) for the final leg.}
    \label{fig:switch_SPD_APD}
\end{figure}

As discussed in Section~\ref{subsec:angle_pd}, the angle of the PD should ideally be adjusted depending on how far away the drone is from the light source at the landing station. As such, we incorporate a motorized PD whose angle can be tuned by the drone. Additionally, we also attach two additional downward PDs on the drone as shown in Figure~\ref{fig:switch_SPD_APD} in an equilateral triangle shape. We noticed that when the measured intensity of the two downward facing PDs exceeds the intensity of the motorized PD, the drone is within a distance from the landing station where an array of downward facing PDs can accurately guide the drone to land. This is because when the incident angle of light on the downward facing PDs is greater than an angled PD when the drone is close to the station as discussed in Section~\ref{subsec:angle_pd}. The exact distance depends on the height of the drone, as we will discuss in Section~\ref{sec:eval_yanchen}. Hence, we use these two additional PDs to determine when to switch from the single motorized PD to an array of PDs for the final leg. We use an array of PDs for the final leg, rather than a single PD, because of the comparatively large drifts and errors arising from drone movements in this close regime, as discussed in Section~\ref{subsec:sensing_configurations}. This method combines both the benefits of a single side facing PD, for guidance at long range, along with an array of downward facing PDs at short range.

The full landing procedure is as follows. First, the drone sweeps $360$ degrees along its yaw ``spinning'' in place, before orienting the motorized PD at the front of the drone in the direction with the greatest intensity. Next, \name sweeps the polar angles of the motorized PD and angles the PD at the angle with the greatest intensity. \name moves the drone forward while increasing the angle of the motorized PD commensurate to the speed at which the light intensity measured by the PD is increasing. Finally, when the intensity of the two downward facing PDs exceeds the motorized PD, \name leverages all three PDs as an array of downward facing PDs (ArPD) to guide the drone. Once the intensity measured by each PD is equal, the drone is directly on top of the light source and descends to finish landing.

%% file: sections/nlos.tex
\section{Exploring NLOS Navigation}
\label{sec:explore_nlos}

\begin{figure}
    \centering
    \includegraphics[width=0.9\linewidth]{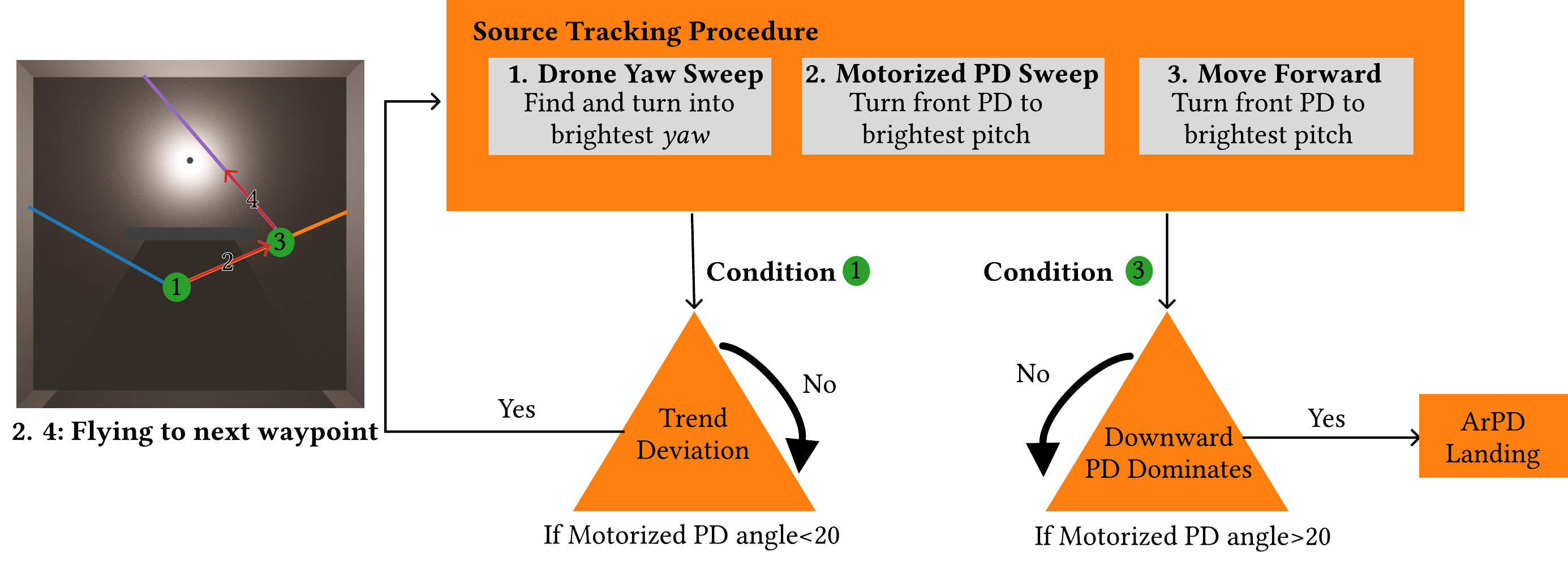}
    \caption{NLOS navigation example process.}
    \label{fig:nlos_nav}
\end{figure}

In this section, we discuss non line of sight (NLOS) landing scenarios where the drone begins at a location that is partially occluded from the light source. Consider the scenario in Figure~\ref{fig:nlos_nav}, where the drone starts from behind a wall. In this case, the drone should 1) navigate to the opening before 2) reorienting and moving to the landing station. This case is commonly found in many indoor scenarios where the drone might need to navigate to a neighboring room or through a door to land. In Section~\ref{sec:nlos_eval} we analyze the performance of our NLOS path planning and discuss scenarios that require future work.

\subsection{Navigating Towards Openings}
\label{subsec:navigate_opening}

In Figure~\ref{fig:nlos_nav}, the starting position of the drone is not in line of sight of the landing station. However, it is still in line of sight of the opening that it should move towards. First, if the drone sweeps $360$ degrees and measures the light intensity, we see that the direction of greatest intensity corresponds to the direction of the opening, where a light path from the lightbulb passes through. Second, we see that as the drone approaches the opening, the angle of the PD that receives the greatest intensity measurements also increases. Both of these observations mimic the same properties as the LOS case, except the opening acts as the light source rather than the landing station. This is because nearby wall, obstacles, and furniture acts as pseudo light sources when light from the IR bulb are diffused and reflected off of them. As such, \textbf{we can guide the drone towards openings using the same method that guides drones to the landing station in the LOS scenario}.

\subsection{Reorienting Towards Landing Station}
\label{subsec:reorienting}

Unlike the LOS scenario (Section~\ref{sec:lightfield_sensing_on_drone}), the drone needs to stop and reorient itself when it reaches the opening. In the LOS case, the drone changes between using the single motorized PD and the array of downward facing PDs when the intensity measured by the downward facing PDs exceeds the single motorized PD, which signifies that the drone is close to the landing station. However, the drone can still be far from the landing station even after reaching the opening, and the drone is likely still not within range where the downward facing PDs would measure more light intensity than the motorized and angled PD. As such, borrowing this stopping criterion from the LOS scenario would likely cause the drone to not stop and crash into the wall.

As shown in Figure~\ref{fig:nlos_nav}, when the drone approaches the opening or the light source at the landing station, the measured light intensities grow at a polynomial rate. In the LOS scenario, this trend continues until the drone arrives near the landing station and can begin using the downward facing PDs in the final leg. In the NLOS setting, when the drone reaches the opening and comes into LOS of the landing station, there is an exponential jump in measured intensities before returning back to a polynomial rate of change. This occurs since the PDs now receive light in the direct path from the lightbulb as well as the the reflected and diffused light from the surrounding obstacles. As such, we can detect this exponential jump or barrier function to stop the drone from continuing forward, before sweeping $360$ degrees around the drone and the motorized PD to reorient itself, just as in the LOS scenario.

%% file: sections/implementation.tex
\section{Implementation}
\label{sec:implementation}

\begin{figure*}[t]
    \centering \includegraphics[width=\linewidth]{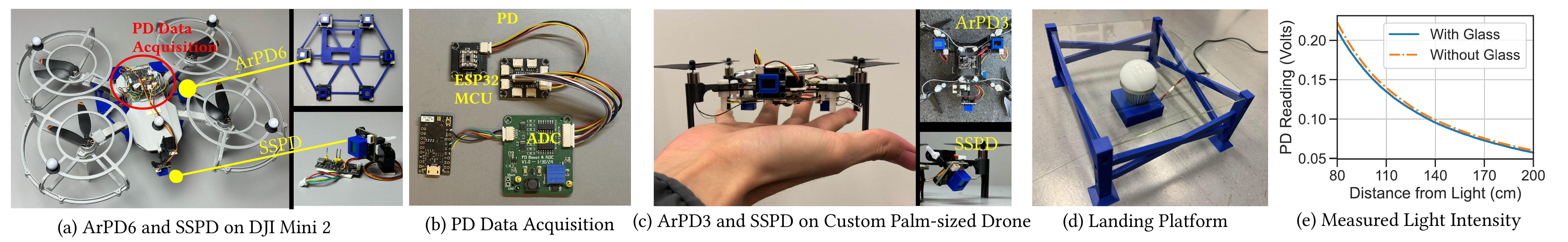}
    \caption{(a, b, c) \name Implementation; (d) landing platform with anti-reflective glass; (e) Measured light intensity vs. distance with and without installing the glass landing platform, showing minimal impact of the anti-reflective glass to the light.}
    \label{fig:sys_implementation}
\end{figure*}

\subsection{Platform for Sensing Light Fields on Drones}
\label{subsec:implementation_ondrone_sensing}

\begin{figure}
    \centering
    \includegraphics[width=0.9\linewidth]{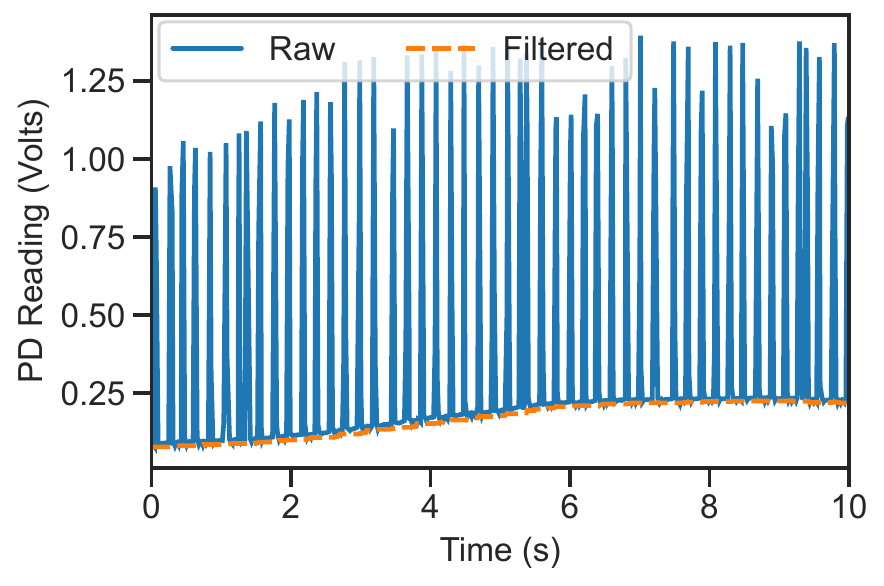}
    \caption{The drone's IR height sensor emits IR pulses that cause large variations in PD measurements since it operates within the same range as our PD (860nm and 960nm). We apply a rolling minimum filter to remove this interference.}
    \label{fig:filtering_optic_flow}
\end{figure}

\textbf{1. Hardware Platform.} Figure~\ref{fig:sys_implementation}c shows our hardware platform, capable of being attached to the underside of a small drone. One Motorized PD is attached at the front of the drone while the other two PDs are installed facing downward. Three PDs are placed 4cm from the center of the drone as equilateral triangle. We used the TI OPT101 PD to implement the sensing platform~\cite{tiopt101pd}.

In our deployments (Section~\ref{sec:eval_yanchen}), we integrate our hardware platform on a DJI Mini 2~\cite{djimini2} and control the drone based on sensed light through Rosetta Drone~\cite{rosettadrone}, an open source Mavlink wrapper that allows us to programmatically control DJI drones through software libraries. We use the DJI Mini 2 purely for demonstration purposes, turning off Mini 2's camera and using only \name's hardware platform for landing guidance.

\textbf{2. Impact of IR Height Sensor.} Drones commonly leverage IR distance sensors to measure its height from the ground. This typically involves emitting and measuring the response of IR pulses, which causes the measurements from the PDs to fluctuate wildly as the laser turns on and off, as shown in Figure~\ref{fig:filtering_optic_flow}. To remove this interference, we take a rolling minimum average to capture PD measurements when the IR laser turns off and interpolate segments where the IR laser turns on.

\subsection{Landing Station and Light Field Generation}
\label{subsec:lightfield_generation}

\begin{figure}[]
  \begin{center}
    \includegraphics[width=0.4\textwidth]{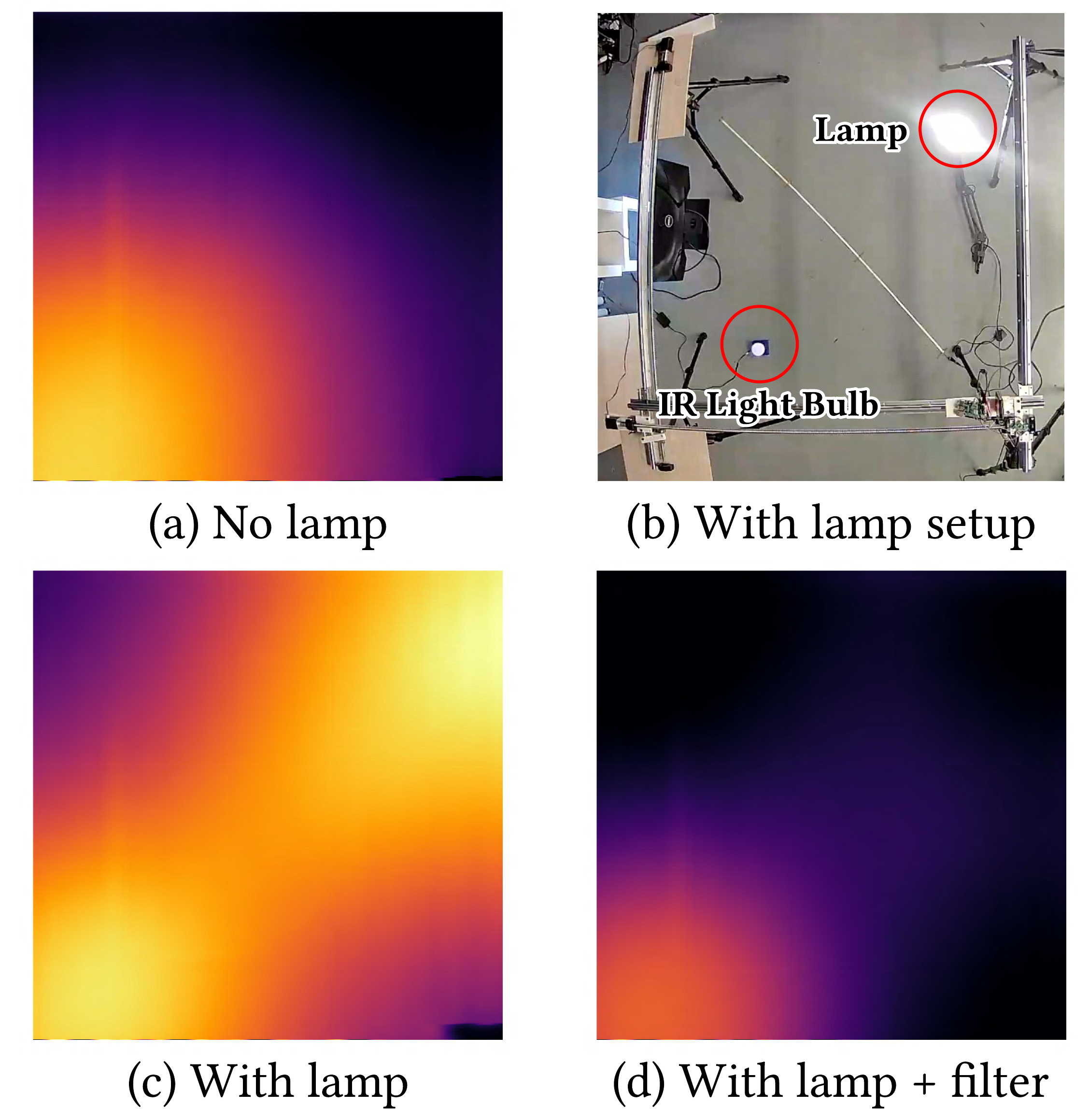}
  \end{center}
  \caption{Effect of external light sources.}
  \label{fig:lightfield_external_lightsources}
\end{figure}

\begin{figure}[]
  \begin{center}
    \includegraphics[width=0.48\textwidth]{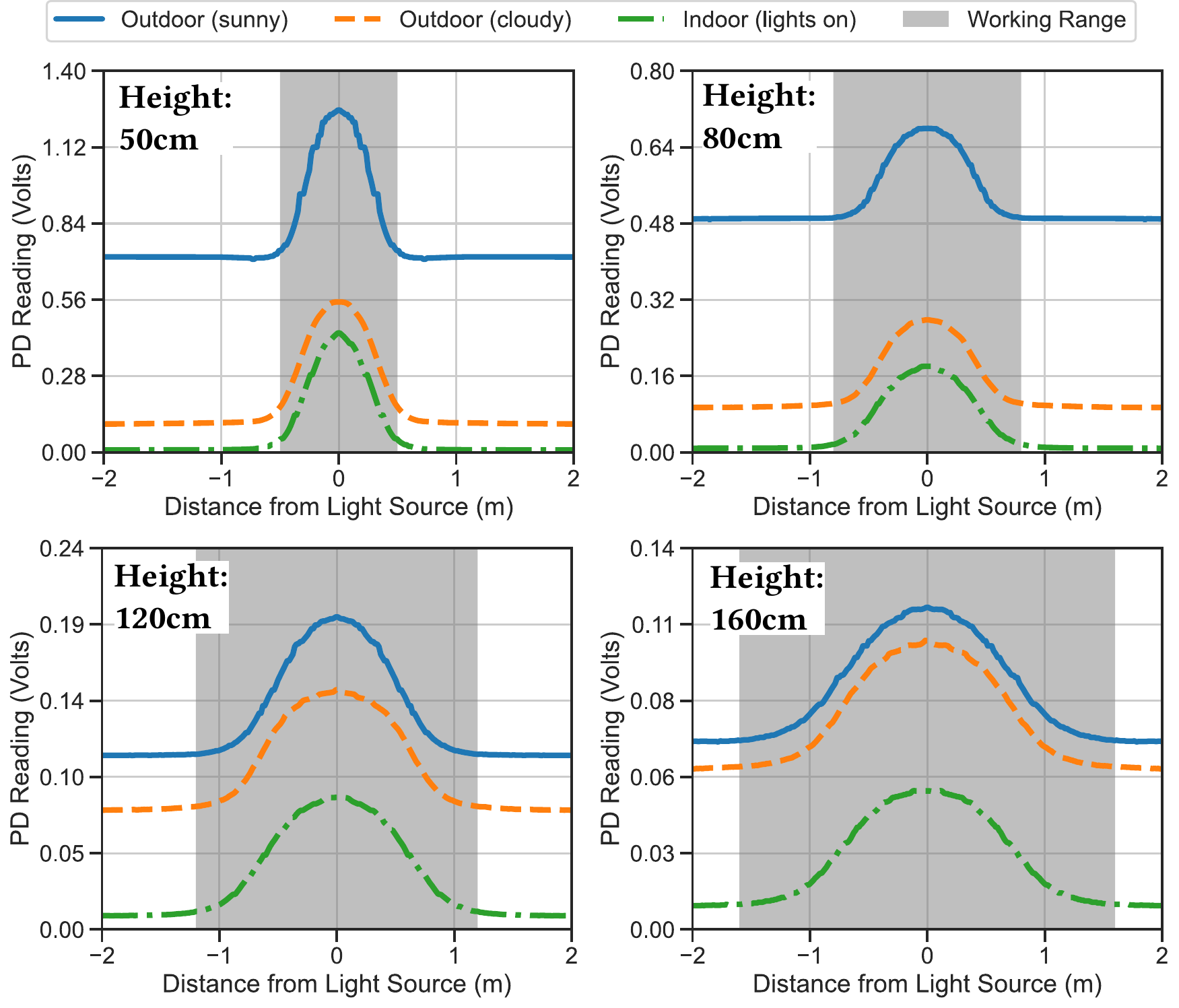}
  \end{center}
  \caption{Effect of external sunlight on noise floor of measured light from PD.}
  \label{fig:lightfield_noise_floor}
\end{figure}

\textbf{1. Landing Station.} To create the landing platform, we center the IR light bulb underneath a 20 by 25 centimeter sheet of clear anti-reflective glass that the drone will land on. We measure the effect of the anti-reflective glass on the radiation pattern of the IR light bulb. Figure~\ref{fig:sys_implementation}c shows the intensity of the light field as a function of distance from the center of the light source, at specific heights above the light bulb, with and without the anti-reflective glass covering the light bulb. We see from our measurements that the glass has minimal effect on the radiation pattern.

\noindent
\textbf{2. External Light Sources.} We also measure the effect of external light sources on the measured IR light field. Figure~\ref{fig:lightfield_external_lightsources} shows our setup. We set a high powered lamp next to the IR light bulb and measure the light field at $1.1$m above the ground (Figure~\ref{fig:lightfield_external_lightsources} upper right). External light sources, like a lamp, can emit high amounts of IR light in addition to visible light. Using a spectrometer, we measured that the IR light bulb emits high energy near $840$nm. We try placing filters on top of our PDs to filter out light frequencies in other bands. The impact of the lamp has almost been completely removed after adding a $940$nm band pass filter (BPF) and an extra $780$nm low pass filter (LPF) (Figure~\ref{fig:lightfield_external_lightsources} lower right).

\noindent
\textbf{3. Outdoors.} If \name is used outdoors, the sun is a major ambient source of IR light. We observed that this causes the PD to saturate. However, applying our $940$nm BPF and $780$nm LPF, shown in Figure~\ref{fig:lightfield_noise_floor}, reduces the noise floor significantly, and we can now observe changes in light readings caused by the IR light bulb. Figure~\ref{fig:lightfield_noise_floor} measures the $X$/$Y$ cross section of our IR light field at various heights and indoors vs. outdoors. In gray, we high light the \textit{operating range}, or the region above the noise level that is not flat. In this region PDs can sense light from the landing station and guide the drone. We see that this range varies depending on the height of the drone from the ground, as well as the orientation of the PDs, as we will discuss in Section~\ref{sec:eval_yanchen}.

We see that the measured light field outdoors is almost identical to the indoor scenario, except with a DC offset caused by the extra light from the sun. Because our method relies on light gradients or changes, rather than absolute intensities to determine the direction to move the drone, this DC offset does not affect the performance of our system. Moreover, the potential \textit{operating range} of \name also remains unaffected between these different lighting conditions. In summary, \textbf{\name remains largely immune to different lighting conditions or external sources of light in the environment}.

%% file: sections/eval.tex
\section{Line of Sight Evaluation}

\label{sec:eval_yanchen}

\begin{figure*}[]
    \centering \includegraphics[width=0.65\linewidth]{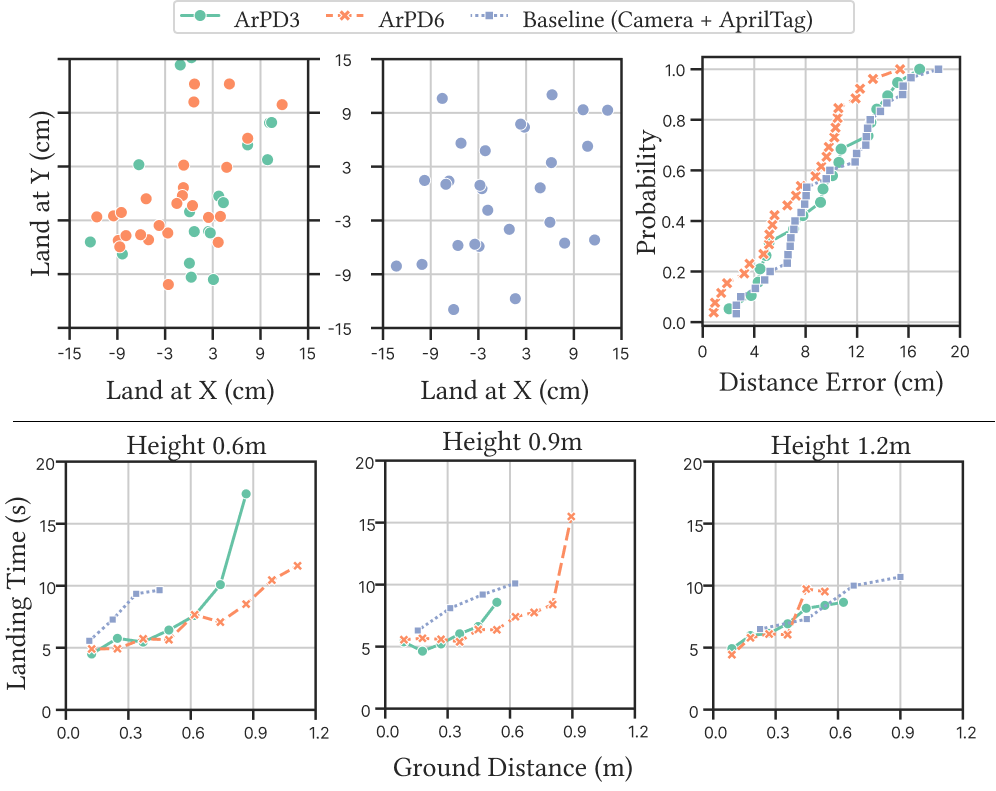}
    \caption{End-to-end landing landing time and accuracy comparison. Top shows scatterplot and CDF for landing locations and error. Bottom shows landing time against ground distance of landing start point at different heights.}
    \label{fig:end_to_end_eval}
\end{figure*}

\begin{figure*}[]
    \centering \includegraphics[width=\linewidth]{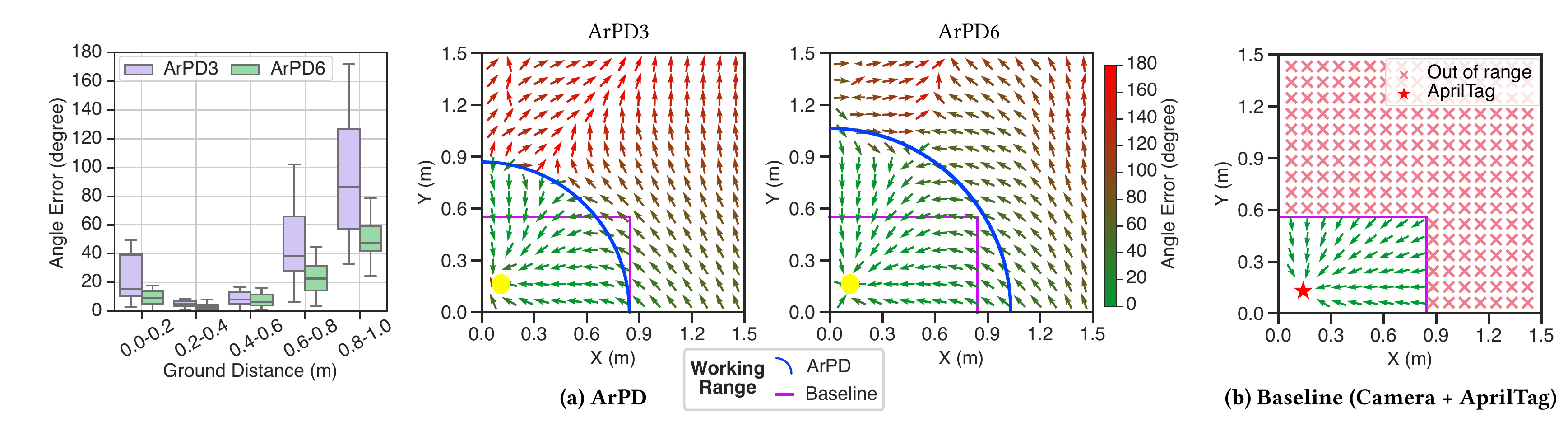}
    \caption{Estimated movement directions at $1$m height in a typical indoor setting for a) ArPD and b) camera-based baseline.}
    \label{fig:eval_story}
\end{figure*}

\begin{figure*}
    \centering
    \includegraphics[width=0.9\linewidth]{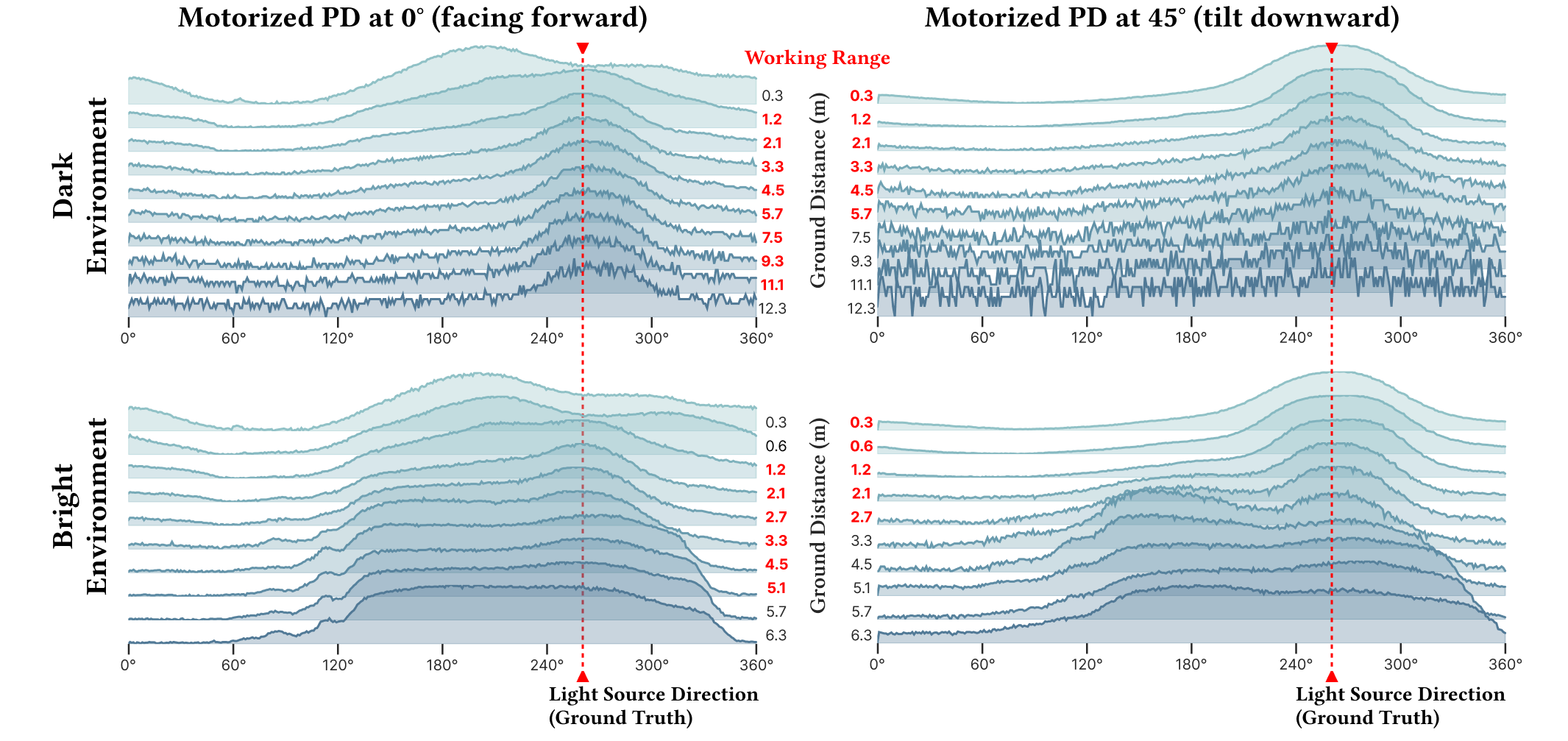}
    \caption{Light intensity measurements at different distances from the landing station sweeping $360$ degrees around the drone's yaw axis. a) a side facing PD ($0$ degrees) and b) a downward facing PD ($90$ degrees). The ranges in which these two PDs can accurately measure the direction of the light source are exactly opposite from one another}
    \label{fig:sweeping_pds}
\end{figure*}

\noindent
\textbf{1. Baselines and Metrics.} Figure~\ref{fig:end_to_end_eval} shows the end-to-end landing time, direction estimation error, and distribution of landing locations when deploying \name in a real indoor LOS setting. Normal room lights were used. We compared against three baselines:

\begin{itemize}
\item Single Motorized PD (MPD)
\item Array of 3 and 6 downward facing PDs (ArPD3 and ArPD6)
\item RGB Camera-based guidance to detect AprilTag markers at the landing station.
\end{itemize}

The CDF of direction estimation errors measures errors within the ``operating range'' where each method is capable of accurately detecting and guiding the drone to the light source. The landing time plots also reflect just this region.

\noindent
\textbf{2. Landing accuracy and time.} The average landing time within the final leg and error of each method within their respective operating range is as follows:
\begin{itemize}
\item ArPD3: 7.1 sec, 9.2cm
\item ArPD6: 7.3 sec, 7.3cm
\item AprilTag: 8.3 sec, 9.4cm
\end{itemize}

Metrics for the motorized PD is not reported because, as we will see next, it is not possible to accurately guide the drone to land during the last leg using a single side angled PD. ArPD3 was able to achieve slightly lower landing error than the baseline camera-based method at a faster speed. Increasing the number of PDs allowed \name to more accurately land the drone.

\noindent
\textbf{3. Operating Range.} The real difference arises when looking at the operating range in which each method can achieve these low errors; outside of these ranges, the drone can no longer reliably detect the location of the landing station. Figure~\ref{fig:sweeping_pds} shows the the raw light intensity measurement sweeping $360$ degrees along the drone's yaw axis, while hovering at $1.2$ meters. We see that the upper and lower bound of the operating range of the side facing and downward facing PD, respectively, is at $0.9$ meters away from the landing station, which is exactly complementary to one another. We observed that \textit{the side facing PD can still detect the direction of the light source up to around $40$ meters}, where it reaches the limit of the resolution of the analog-to-digital converter (ADC).

Figure~\ref{fig:eval_story} shows the estimated direction of greatest light intensity using an array of 3 and 6 downward facing PDs (ArPD3 and ArPD6) at various locations around the landing station with the drone hovering at 1 meter. We take the ArPD6 and ArPD3 measurements in a real indoor deployment on a DJI Mini 2 (ArPD6, Figure~\ref{fig:sys_implementation}(a)) and custom designed palm-sized drone (ArPD3, Figure~\ref{fig:sys_implementation}(c)), described in Sections~\ref{sec:implementation} and~\ref{sec:eval_yanchen}. This result is compared against a vision landing baseline implemented on the DJI Mini2 drone using its RGB camera and AprilTag. The baseline method has a rectangular operating range due to the rectangular nature of images. For our light-based approach, we see that adding more PDs increases the working range of the downward facing PDs. However, leveraging 3 PDs can still give a greater range than using camera-based methods. Moreover, \textit{\name combines both the motorized side facing PD and the array of downward facing PDs to get the best of both range and landing accuracy}, as discussed in Section~\ref{subsec:proposed_motorized_spd}. At a typical height of $1$m, \name can reliably guide the drone within $11.1$ meters to the landing station, while ArPD6 fails past $0.9$ meters and leveraging camera-based AprilTag markers fails past $0.6$ meters.

\noindent
\textbf{4. Power Consumption.} \name leverages 3 PDs, which adds $\mu W$ level power consumption per PD. This addition is negligible on drones, whose motors consume Watt-level power. After outfitting the DJI Mini 2 with \name, there was no noticeable change to the drone's battery life.

%% file: sections/eval_nlos.tex
\section{Non Line of Sight Evaluation}
\label{sec:nlos_eval}

\begin{figure*}[]
    \centering \includegraphics[width=.83\linewidth]{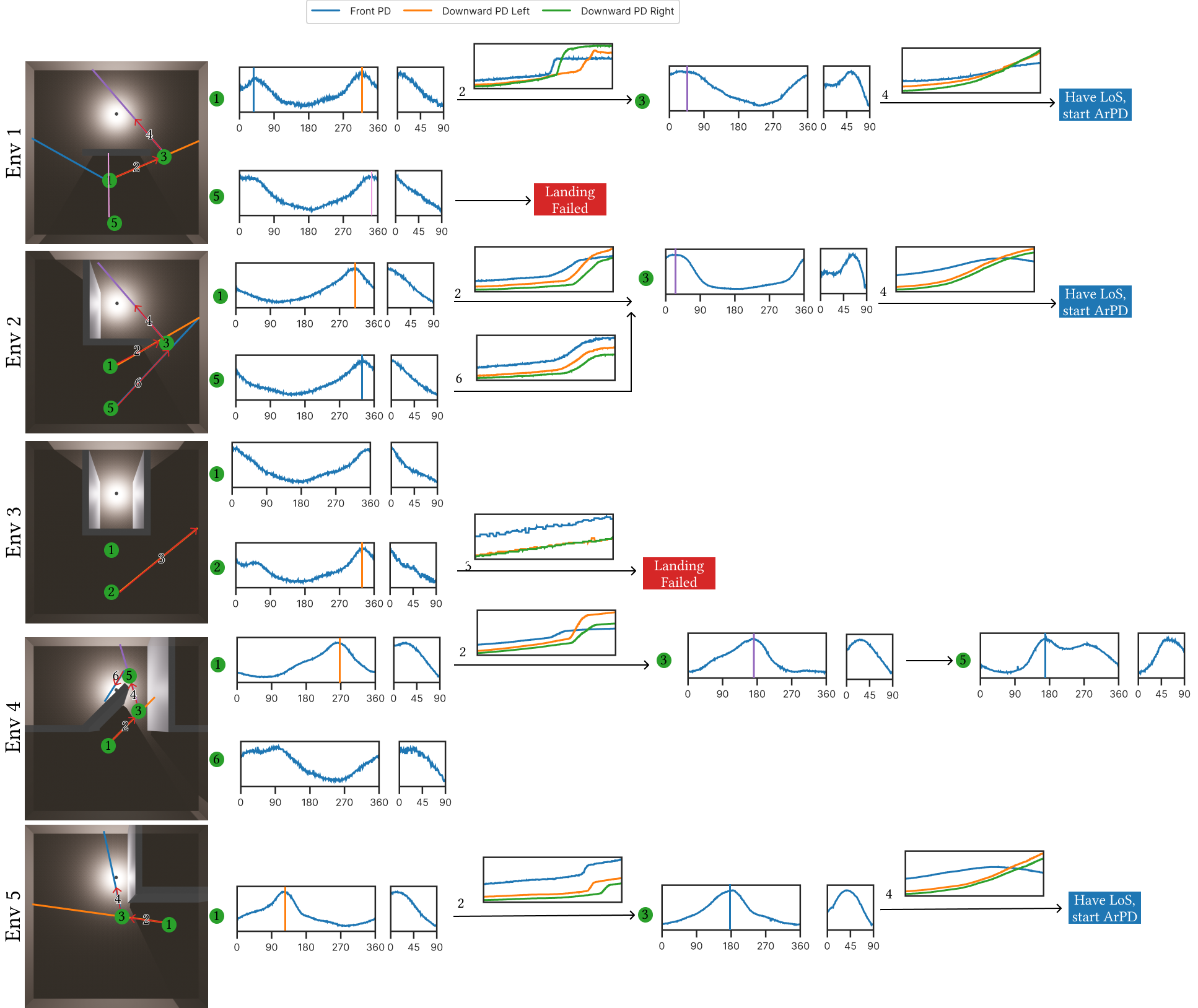}
    \caption{Evaluation and trajectories for drone guidance and landing for common indoor NLOS scenarios.}
    \label{fig:nlos_eval}
\end{figure*}

Figure~\ref{fig:nlos_eval} shows the NLOS scenarios, including mockups for clarity, that we deployed \name into, as well as example trajectories and PD intensities observed. These scenarios typically involve the drone seeking out the landing station behind an obstacle (env 3), turning a corner (env 5), or through a door (env 4). For each of these environments, we deployed \name ten times and show example trajectories of both successful and unsuccessful runs for analysis. For environments 2, 4, and 5 \name was able to guide the drone successfully in all attempts, encompassing common scenarios such as moving through doors (env 4), turning corners (env 5), and entering a partially open space (env 2). 

For environment 1, most of our runs began around point 1, near the wall and closer to the landing station and were successful. The run that began at point 5 failed to detect when the drone entered the opening and became LOS of the landing station because it was further than 15 feet out from the landing station (From Section~\ref{sec:eval_yanchen}, the max range we measured is approximately 11 feet). This meant that \name was not able to reliably detect the light from the landing station, nor detect the trend deviation (Section~\ref{subsec:reorienting}) in light intensity to stop and reorient the drone. In environment 3, the drone was completely non line of sight of any light path emitting from the landing station. Although \name could not guide the drone in scenarios where all light was completely non line of sight or too far from the drone to sense, \name can still operate in many common partially non line of sight scenarios within $11$ meters of the landing station.

%% file: sections/futurework.tex
\section{Discussion}\label{sec:futurework}

\noindent
\textbf{Optimizing Methods and Algorithms}. The landing methods proposed in this work guide the drone directly above the landing station, before descending, essentially estimating 2D light gradients. In future work, we plan to explore more array layouts and methods that allow the drone to travel in three dimensions, enabling the system to descend while traveling to the landing station. This could potentially reduce landing errors and landing time by removing the final descent once the drone lands at the center of the landing pad.

\noindent
\textbf{Complete Non Line of Sight.} As mentioned in Section~\ref{sec:nlos_eval}, \name can only guide the drone to the landing station in partially non line of sight scenarios where the drone is still in line of sight of a light path emitted from the light source and within reasonable distance to measure this secondary. In cases where the drone is not in line of sight of a light path, or this path is too weak to measure, the drone needs to rely on other methods. We plan to explore scenarios and methods that involve deploying multiple light sources or tags to guide the drone across multiple barriers or rooms.